\documentclass{article}

\usepackage[final, nonatbib]{include/neurips_data_2021}

\usepackage[utf8]{inputenc} %
\usepackage[T1]{fontenc}    %
\usepackage{hyperref}       %
\usepackage{url}            %
\usepackage{booktabs}       %
\usepackage{amsfonts}       %
\usepackage{nicefrac}       %
\usepackage{microtype}      %
\usepackage[numbers]{natbib}

\usepackage[noabbrev, capitalise]{cleveref}
\usepackage{graphicx}
\usepackage{multicol}
\usepackage{multirow}
\usepackage{verbatim}
\usepackage[dvipsnames]{xcolor}
\usepackage{caption}
\usepackage{subcaption}%
\usepackage{float}
\usepackage{wrapfig}
\usepackage{listings}
\usepackage{pifont}
\usepackage{enumitem}
\usepackage{tikz}
\usepackage[frozencache,cachedir=.]{minted}%

\newcounter{commentCounter}
\newif\iftrvar
\trvartrue
\iftrvar
\newcommand{\tim}[1]{{\small \color{red} \refstepcounter{commentCounter}\textsf{[TR]$_{\arabic{commentCounter}}$:{#1}}}}
\newcommand{\mika}[1]{{\small \color{blue} \refstepcounter{commentCounter}\textsf{[MS]$_{\arabic{commentCounter}}$:{#1}}}}
\newcommand{\ed}[1]{{\small \color{magenta} \refstepcounter{commentCounter}\textsf{[ETG]$_{\arabic{commentCounter}}$:{#1}}}}
\newcommand{\rob}[1]{{\small \color{darkgreen} \refstepcounter{commentCounter}\textsf{[RK]$_{\arabic{commentCounter}}$:{#1}}}}

\else
\newcommand{\mika}[1]{}
\newcommand{\tim}[1]{}
\newcommand{\ed}[1]{}
\newcommand{\rob}[1]{}

\fi

\definecolor{lightgray}{rgb}{.9,.9,.9}
\definecolor{darkgray}{rgb}{.4,.4,.4}
\definecolor{purple}{rgb}{0.65, 0.12, 0.82}
\definecolor{darkgreen}{rgb}{0, 0.365, 0}

\lstdefinelanguage{des}{
  keywords={MONSTER, MAZE, MAP, ENDMAP, LOOP, TRAP, OBJECT, GOLD, GEOMETRY, FLAGS, SHUFFLE, TERRAIN, REPLACE_TERRAIN, STAIR, REGION, BRANCH, IF, ELSE, LEVEL, ROOM, SUBROOM, ENTITY, SINK, FOUNTAIN},
  keywordstyle=\color{blue}\bfseries,
  keywords=[2]{\$river, \$place , \$monster, \$roll, \$mon_letters, \$mon_names, \$mon_index, \$object, \$variable_name, \$center, \$apple_location},
  keywordstyle=[2]\color{purple}\bfseries,
  identifierstyle=\color{black},
  sensitive=false,
  comment=[l]{\#},
  morecomment=[s]{/*}{*/},
  commentstyle=\color{darkgreen}\ttfamily,
  stringstyle=\color{red}\ttfamily,
  morestring=[b]',
  morestring=[b]"
}

\newcommand{\NLE}{\texttt{NLE}}
\newcommand{\desfile}{\texttt{des-file}}

\title{MiniHack the Planet: A Sandbox for Open-Ended Reinforcement Learning Research}

\newcommand{\fair}{\texttt{\large +}}
\newcommand{\oxford}{\texttt{\large \#}}
\newcommand{\ucl}{\texttt{\large !}}

\author{%
    \textbf{Mikayel Samvelyan}$^{\fair{} \ucl{}}$
    \textbf{Robert Kirk}$^\ucl{}$
    \textbf{Vitaly Kurin}$^\oxford{}$
    \textbf{Jack Parker-Holder}$^\oxford{}$
    \textbf{Minqi Jiang}$^{\fair{} \ucl{}}$\\
    \textbf{Eric Hambro}$^\fair{}$
    \textbf{Fabio Petroni}$^\fair{}$
    \textbf{Heinrich K{\"u}ttler}$^\fair{}$
    \textbf{Edward Grefenstette}$^{\fair{} \ucl{}}$
    \textbf{Tim Rockt{\"a}schel}$^{\fair{} \ucl{}}$\\[1em]
    $^\fair{}$Facebook AI Research ${}^\ucl{}$University College London 
    ${}^\oxford{}$University of Oxford \\[0.5em]
    \texttt{\{samvelyan,rockt\}@fb.com}
}

\begin{document}

\maketitle

\begin{abstract}
Progress in deep reinforcement learning (RL) is heavily driven by the availability of challenging benchmarks used for training agents.
However, benchmarks that are widely adopted by the community are not explicitly designed for evaluating specific capabilities of RL methods. 
While there exist environments for assessing particular open problems in RL (such as exploration, transfer learning, unsupervised environment design, or even language-assisted RL), it is generally difficult to extend these to richer, more complex environments once research goes beyond proof-of-concept results. 
We present MiniHack, a powerful sandbox framework for easily designing novel RL environments.\footnote{Code available at \url{https://github.com/facebookresearch/minihack}} 
MiniHack is a one-stop shop for RL experiments with environments ranging from small rooms to complex, procedurally generated worlds.
By leveraging the full set of entities and environment dynamics from NetHack, one of the richest grid-based video games, MiniHack allows designing custom RL testbeds that are fast and convenient to use. 
With this sandbox framework, novel environments can be designed easily, either using a human-readable description language or a simple Python interface. 
In addition to a variety of RL tasks and baselines, MiniHack can wrap existing RL benchmarks and provide ways to seamlessly add additional complexity.

\end{abstract}

\section{Introduction}
Advancing deep reinforcement learning~\citep[RL,][]{suttonbarto} methods goes hand in hand with developing challenging benchmarks for evaluating these methods.
In particular, simulation environments like the Arcade Learning Environment \citep[ALE,][]{bellemare2016unifying} and the MuJoCo physics simulator \cite{todorov2012mujoco} have driven progress in model-free RL and continuous control respectively. However, after several years of sustained improvement, results in these environments have started to reach superhuman performance \cite{ecoffet2019go, pbt_mujoco, Agent57} while many open problems in RL remain~\cite{dulac_arnold2020empirical, ibarz2021lessons, Hill2020Environmental}. To make further progress, novel challenging RL environments and testbeds are needed.

On one hand, there are popular RL environments such as Atari~\citep{ale}, StarCraft~II~\citep{vinyals2017starcraft}, DotA~2~\citep{openai2019dota}, Procgen~\cite{cobbe2019procgen}, Obstacle Tower~\cite{juliani2019obstacle} and NetHack~\citep{kuttler2020nethack} that consist of entire games, but lack the ability to test specific components or open problems of RL methods in well-controlled proof-of-concept test cases. 
On the other hand, small-scale tightly controlled RL environments such as MiniGrid~\citep{gym_minigrid}, DeepMind Lab~\cite{BeattieLTWWKLGV16}, Alchemy~\cite{wang2021alchemy}, MetaWorld~\cite{yu2019meta}, and bsuite~\cite{osband2020bsuite} have emerged that enable researchers to prototype their RL methods as well as to create custom environments to test specific open research problems (such as exploration, credit assignment, and memory) in isolation.
However, once specific research hypotheses are verified in these controllable simplified environments, RL practitioners find themselves between a rock and a hard place. Systematically extending such environments and gradually dropping simplifying assumptions can require arduous engineering and excessive time commitment, while opting for more challenging benchmarks \cite[e.g.][]{kuttler2020nethack} often deprives researchers of a controllable path for assessing subsequent hypotheses. While frameworks like PyVGDL~\cite{schaul_video_2013}, GVGAI~\cite{GVGAI}, and Griddly~\cite{griddly} can be used to design custom testbeds, creating complex environments with rich entities and environment dynamics would still require substantial engineering effort as complex environment interactions would have to be designed from scratch. Thus, there is a gap in terms of a framework that allows one to easily specify a suite of rich, gradually more difficult tasks, while also providing a large set of entities with complex environment interactions ready to be used to implement these tasks.

\begin{figure}
\foreach \x in {0,1,2,3,4,5,6,7}
{ 
    \includegraphics[width=0.1132\textwidth]{figures/screens/mzwk_\x.png}
}\\
\includegraphics[width=0.49\textwidth]{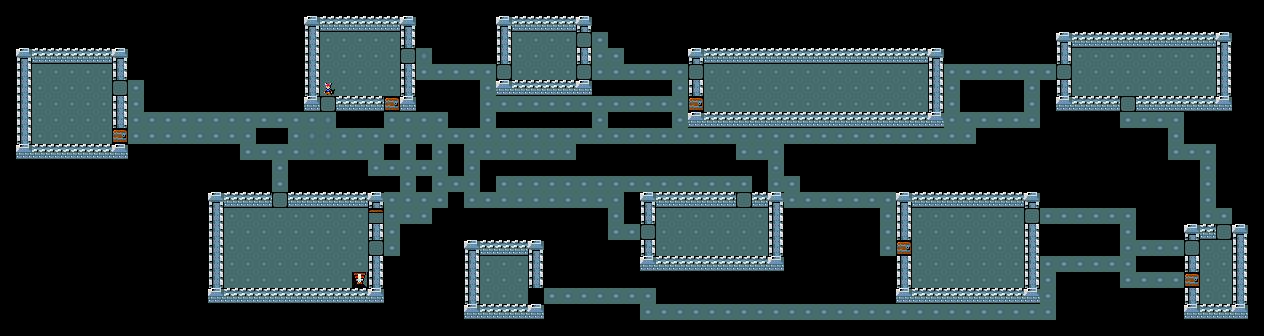}~
\includegraphics[width=0.49\textwidth]{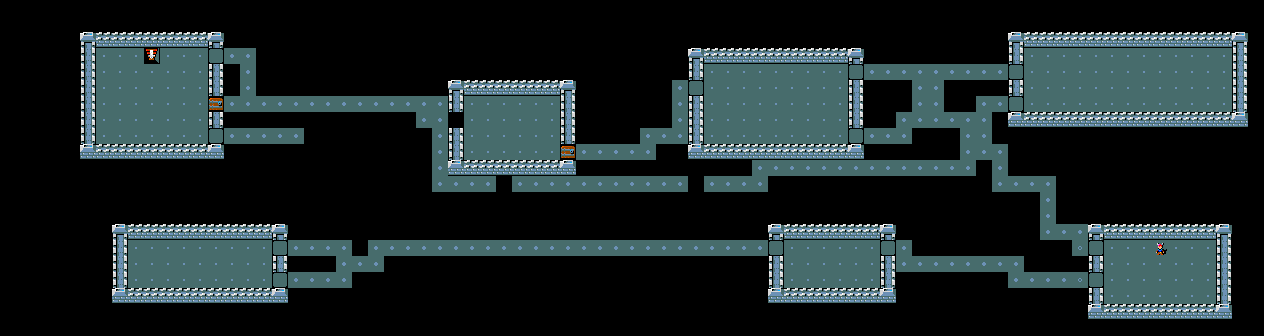}\\  
\foreach \x in {0,1,2,3,4,5,6,7}
{ 
    \includegraphics[width=0.115\textwidth]{figures/screens/bb_\x.png}
}
\caption{Examples of procedurally generated environments using the \desfile{} format. \textbf{(Top)}:  \texttt{MAZEWALK} command applied on a \texttt{15x15} grid, \textbf{(Middle)} corridors generated via \texttt{RANDOM\_CORRIDOR}, \textbf{(Bottom)}: environments generated using the code snippet from \cref{code:des_procgen}.}
\label{fig:procgen}
\vspace{-3mm}
\end{figure}

To fill this gap, we present MiniHack, a sandbox framework for easily designing novel RL environments and enriching existing ones.
At the core of MiniHack are description files for defining procedurally generated worlds via the powerful domain-specific language (DSL) of the game of NetHack~\cite{raymond2020guide}.
The full game of NetHack, arguably the richest gridworld benchmark in RL~\cite{kuttler2020nethack}, is not suitable for answering specific research questions in isolation.
However, NetHack's DSL allows MiniHack to tap into the richness of the game with its hundreds of pre-implemented entities and the complex interaction mechanics between them \cite{nhwiki}.
Furthermore, this DSL is flexible enough to easily build a wide range of testbeds, creating rich and diverse custom environments using only a few lines of human-readable code (see examples in \Cref{fig:procgen}). 
Once written, either directly or using a convenient Python interface, MiniHack compiles the provided description files and wraps them as standard Gym environments \cite{DBLP:journals/corr/BrockmanCPSSTZ16}.

The clear taxonomy of increasingly difficult tasks, the availability of multi-modal observations (symbolic, pixel-based, and textual), its speed and ease of use, make MiniHack an appealing framework for a variety of different RL problems. 
In particular, MiniHack could be used to make progress in areas such as unsupervised skill discovery, unsupervised environment design, transfer learning, and language-assisted RL. 
In addition to a broad range of environments that can easily be designed in the MiniHack framework, we also provide examples on how to import other popular RL benchmarks, such as MiniGrid \cite{gym_minigrid} or Boxoban \cite{boxobanlevels}, to the MiniHack planet. Once ported, these environments can easily be extended by adding several layers of complexity from NetHack (e.g. monsters, objects, dungeon features, stochastic environment dynamics, etc) with only a few lines of code. 

In order to get started with MiniHack environments, we provide a variety of baselines using frameworks such as TorchBeast \cite{torchbeast2019} and RLlib \cite{pmlr-v80-liang18b},
as well as best practices for benchmarking (see \cref{appendix:eval_methodology}).
Furthermore, we demonstrate how it is possible to use MiniHack for unsupervised environment design, with a demonstration of the recently proposed PAIRED algorithm \cite{paired}. Lastly, we provide baseline learning curves in Weights\&Biases format\footnote{\url{https://wandb.ai/minihack}} for all of our experiments and a detailed documentation of the framework.\footnote{\url{https://minihack.readthedocs.io}}

In summary, this paper makes the following core contributions: (i) we present MiniHack, a sandbox RL framework that makes it easy for users to create new complex environments, (ii) we release a diverse suite of existing tasks, making it possible to test a variety of components of RL algorithms, with a wide range of complexity, (iii) we showcase MiniHack's ability to port existing gridworld environments and easily enrich them with additional challenges using concepts from NetHack, and (iv) we provide a set of baseline agents for testing a wide range of RL agent capabilities that are suitable for a variety of computational budgets.

\section{Background: NetHack and the NetHack Learning Environment}\label{sec:Background}

The NetHack Learning Environment~\citep[\NLE{},][]{kuttler2020nethack} is a Gym interface \cite{DBLP:journals/corr/BrockmanCPSSTZ16} to the game of NetHack \citep{raymond2020guide}.
NetHack is among the oldest and most popular terminal-based games. In NetHack, players find themselves in randomly generated dungeons where they have to descend to the bottom of over 50 procedurally generated levels, retrieving a special object and thereafter escape the dungeon the way they came, overcoming five difficult final levels. Actions are taken in a turn-based fashion, and the game has many stochastic events (e.g. when attacking monsters).
Despite the visual simplicity, NetHack is widely considered as one of the hardest games in history~\cite{15hardestVG}. It often takes years for a human player to win the game for the first time despite consulting external knowledge sources, such as the NetHack Wiki~\citep{nhwiki}. The dynamics of the game require players to explore the dungeon, manage their resources, and learn about the many entities and their game mechanics.
The full game of NetHack is beyond the capabilities of modern RL approaches~\cite{kuttler2020nethack}.

\NLE{}, which focuses on the full game of NetHack using the game's existing mechanisms for procedurally generating levels and dungeon topologies, makes it difficult for practitioners to answer specific research questions in isolation. 
In contrast, with MiniHack we present an extendable and rich sandbox framework for defining a variety of custom tasks while making use of NetHack's game assets and complex environment dynamics.
\section{MiniHack}

MiniHack is a powerful sandbox framework for easily designing novel RL environments. 
It not only provides a diverse suite of challenging tasks but is primarily built for easily designing new ones.
The motivation behind MiniHack is to be able to perform RL experiments in a controlled setting while being able to increasingly scale the difficulty and complexity of the tasks by removing simplifying assumptions.
To this end, MiniHack leverages the description file (\desfile{}) format of NetHack and its level compiler (see \Cref{sec:des}), thereby enabling the creation of many challenging and diverse environments (see \Cref{sec:Tasks}).

\subsection{\desfile{} format: A Domain Specific Language for Designing Environments}
\label{sec:des}

The \desfile{} format \cite{des-file} is a domain-specific language created by the developers of NetHack for describing the levels of the game. \texttt{des-files} are human-readable specifications of levels: distributions of grid layouts together with monsters, objects on the floor, environment features (e.g. walls, water, lava), etc. All of the levels in the full game of NetHack have pre-defined \texttt{des-files}. The \texttt{des-files} are compiled into binary using the NetHack level compiler, and MiniHack maps them to Gym environments.

Levels defined via \desfile{} can be fairly rich, as the underlying programming language has support for variables, loops, conditional statements, as well as probability distributions.
Crucially, it supports underspecified statements, such as generating a random monster or an object at a random location on the map.
Furthermore, it features commands that procedurally generate diverse grid layouts in a single line.
For example, the \texttt{MAZEWALK} command generates complex random mazes (see \cref{fig:procgen}~\textbf{Top}), while the \texttt{RANDOM\_CORRIDORS} command connects all of the rooms in the dungeon level using procedurally generated corridors (see \cref{fig:procgen}~\textbf{Middle}).
\cref{code:des_procgen} presents a \desfile{} code snippet that procedurally generates diverse environment instances on a \texttt{10x10} grid, as presented in \cref{fig:procgen}~\textbf{Bottom}.

\begin{figure}[t]
     \centering
     \begin{minipage}[t]{0.55\textwidth}
\begin{lstlisting}[basicstyle=\footnotesize\ttfamily]{numbers=none}
$river=TERRAIN:{'L','W','I'}
SHUFFLE:$river
LOOP [2] {
  TERRAIN:randline (0,0),(10,10),5,$river[0]
  MONSTER:random,random
}
REPLACE_TERRAIN:(0,0,10,10),'.','T',5%
STAIR:random,down
\end{lstlisting}
\caption{A sample code snippet in \desfile{} format language. The \texttt{\textcolor{purple}{\$river}} variable is used to sample a terrain feature (\textcolor{red}{`\texttt{L}`} for lava, \textcolor{red}{`\texttt{W}`} for water and \textcolor{red}{`\texttt{I}`} for ice). The \textcolor{blue}{\texttt{LOOP}} block draws two rivers via the \texttt{randline} command and places two random monsters at random locations. The \texttt{\textcolor{blue}{REPLACE\_TERRAIN}} commands replaces 5\% of floors (\textcolor{red}{`\texttt{.}`}) with trees (\textcolor{red}{`\texttt{T}`}). A stair down is added at random locations. \label{code:des_procgen}}
     \end{minipage}
     \hfill
     \begin{minipage}[t]{0.4\textwidth}
\begin{lstlisting}[basicstyle=\scriptsize\ttfamily, numbers=left,xleftmargin=0.4cm]
MAZE:"simple_maze",' '
GEOMETRY:center,center
MAP
  --- --- ---  
  |.| |.| |.|  
---S---S---S---
|.......+.+...|
---+-----.-----
|.......+.+...|
---S---S---S---
  |.| |.| |.|  
  --- --- ---  
ENDMAP
LOOP [5] {
	OBJECT:'%',random
	TRAP:random,random
}
[10%]: GOLD: 100,random
MONSTER:('B',"bat"),(3,3)
\end{lstlisting}
\caption{A \desfile{} example for a simple NetHack level.\label{fig:Maze_Des_Example}}   
     \end{minipage}
\end{figure}

\cref{fig:Maze_Des_Example} shows a \desfile{} for a level with fixed, pre-defined map layout (lines 3-13). Here, the `\texttt{.}`, `\texttt{+}`, and `\texttt{S}` characters denote grid cells for floor, closed door, and secret door, respectively, while `\texttt{|}` and `\texttt{-}` denote walls. The loop block (lines 14-17) places five random comestibles (`\texttt{\%}`) and random traps at random positions. Line 18 adds 100 golds at a random location with 10\% probability. Line 19 adds a bat at a fixed location. These examples only provide a glimpse into the variety of levels that can be generated (see \cref{appendix:des_file} for more examples and details on the \desfile{} format). 

\subsection{MiniHack Environments}

\begin{figure}[t]
     \centering
     \begin{subfigure}[b]{0.18\textwidth}
         \centering
         \includegraphics[height=2.5cm]{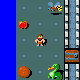}
         \caption{Pixels}
         \label{fig:obs_gui}
     \end{subfigure}
     \hfill
     \begin{subfigure}[b]{0.18\textwidth}
         \centering
         \includegraphics[height=2.5cm]{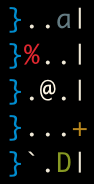}
         \caption{Symbols}
         \label{fig:obs_symb}
     \end{subfigure}
     \hfill
     \begin{subfigure}[b]{0.55\textwidth}
     \footnotesize
     \centering
        \begin{tabular}{|c|c|c|c|c|}
        \hline
        water	&floor	& floor&	killer bee&	wall \\ \hline
    	water	&an apple	& floor&	floor&	wall \\ \hline
    	water	&floor&	\textbf{agent}&	floor&	wall \\ \hline
    	water	&floor&	floor &	floor&	closed door \\ \hline
    	water	&a boulder&	floor&	green dragon&	wall \\ \hline
\end{tabular}
\caption{Textual descriptions}
        \label{fig:obs_text}
     \end{subfigure}
        \caption{Different forms of agent-centred observations of the grid of the map in MiniHack.}
        \label{fig:obs}
\end{figure}

By tapping into the richness of the game of NetHack, MiniHack environments can make use of a large set of pre-existing assets.
One can add one of more than 580 possible monster types, each of which has unique characteristics such as attack distance and type; health points; resistance against certain attacks; and special abilities such as changing shape, moving through walls, and teleporting. 
Practitioners can also choose from 450 items in the game, including various types of weapons, armour, tools, wands, scrolls, spellbooks, comestibles, potions, and more. 
These items can be used by the agent as well as monsters. 

\paragraph{Observations.} MiniHack supports several forms of observations, including global or agent-centred viewpoints (or both) of the grid (such as entity ids, characters, and colours), as well as textual messages, player statistics and inventory information \cite{kuttler2020nethack}. 
In addition to existing observations in \NLE{}, MiniHack also supports pixel-based observations, as well as text descriptions for all entities on the map (see \cref{fig:obs}).

\paragraph{Action Space.} NetHack has a large, structured and context-sensitive action space \cite{raymond2020guide}. We give practitioners an easy way to restrict the action space in order to promote targeted skill discovery. %
For example, navigation tasks mostly require movement commands, and occasionally, kicking doors, searching or eating. Skill acquisition tasks, on the other hand, require interactions with objects, e.g. managing the inventory, casting spells, zapping wands, reading scrolls, eating comestibles, quaffing potions, etc. 75 actions are used in these tasks. 
A large number of actions and their nontrivial interactions with game objects offer additional opportunities for designing rich MiniHack tasks.
For example, a towel can be used as a blindfold (for protection from monsters that harm with their gaze), for wiping off slippery fingers (e.g. after eating deep-fried food from a tin), or even serve as a weapon when wet (which can be achieved by dipping the towel into water).

\paragraph{Reward.} Reward functions in MiniHack can easily be configured. 
Our \texttt{RewardManager} provides a convenient way to specify one or more events that can provide different (positive or negative) rewards, and control which subsets of events are sufficient or required for episode termination (see \cref{appedix:goal_generator} for further details).

\subsection{Interface}\label{sec:Interface}

\begin{wrapfigure}{r}{.42\textwidth}
\vspace{-6mm}
\footnotesize
\begin{minted}
{python}
# Define the labyrinth as a string
grid = """
--------------------
|.......|.|........|
|.-----.|.|.-----|.|
|.|...|.|.|......|.|
|.|.|.|.|.|-----.|.|
|.|.|...|....|.|.|.|
|.|.--------.|.|.|.|
|.|..........|...|.|
|.|--------------|.|
|..................|
--------------------
"""
# Define a level generator
level = LevelGenerator(map=grid)
level.set_start_pos((9, 1))
# Add wand of death and apple
level.add_object("death", "/")
level.add_object("apple",
    place=(14, 5))
# Add a Minotaur at fixed position
level.add_monster(name="minotaur", 
    place=(14, 6), args=("asleep",))

# Define the goal
reward_mngr = RewardManager()
reward_mngr.add_eat_event("apple")

# Declare task a Gym environment
env = gym.make(
    "MiniHack-Skill-Custom-v0", 
    des_file=level.get_des(),
    reward_manager=reward_mngr)
\end{minted}
\vspace{-0.2cm}
\caption{A sample code snippet for creating a custom MiniHack task using the \texttt{LevelGenerator} and \texttt{RewardManager}.\label{code:python_short}}
\vspace{-1.5cm}
\end{wrapfigure}

MiniHack uses the popular Gym interface \cite{DBLP:journals/corr/BrockmanCPSSTZ16} for the interactions between the agent and the environment. 
One way to implement MiniHack Gym environments is to write the description file in the human-readable \desfile{} format and then pass it directly to MiniHack (see \cref{code:des_file} in \cref{appendix:interface}). 
However, users might find it more convenient to construct the environment directly in Python. Our \texttt{LevelGenerator} allows users to do this by providing the functionality to add monsters, objects, environment features, etc.   
\cref{code:python_short} presents an example code snippet of this process. Here, the agent starts near the entrance of a labyrinth and needs to reach its centre to eat the apple. A Minotaur, which is a powerful monster capable of instantly killing the agent in melee combat, is placed deep inside the labyrinth. There is a wand of death placed in a random location in the labyrinth. The agent needs to pick the wand up, and upon seeing the Minotaur, zap it in the direction of the monster. Once the Minotaur is killed, the agent would be able to reach the centre of the labyrinth and eat the apple to complete the task. The \texttt{RewardManager} is used to specify the goal that needs to be completed (eating an apple). %
Our \texttt{LevelGenerator} and \texttt{RewardManager} are described in more detail in \cref{appendix:interface}.

\subsection{Tasks: A World of Possibilities}\label{sec:Tasks}

We release a collection of example environments that can be used to test various capabilities of RL agents, as well as serve as building blocks for researchers wishing to develop their own environments. All these environments are built using the interface described in \cref{sec:Interface}, which demonstrates the flexibility and power of MiniHack for designing new environments.

\begin{figure}[!t]
  \begin{tabular}[b]{lll}
    \begin{tabular}[b]{c}
      \begin{subfigure}[b]{0.29\columnwidth}
        \includegraphics[width=.8\textwidth]{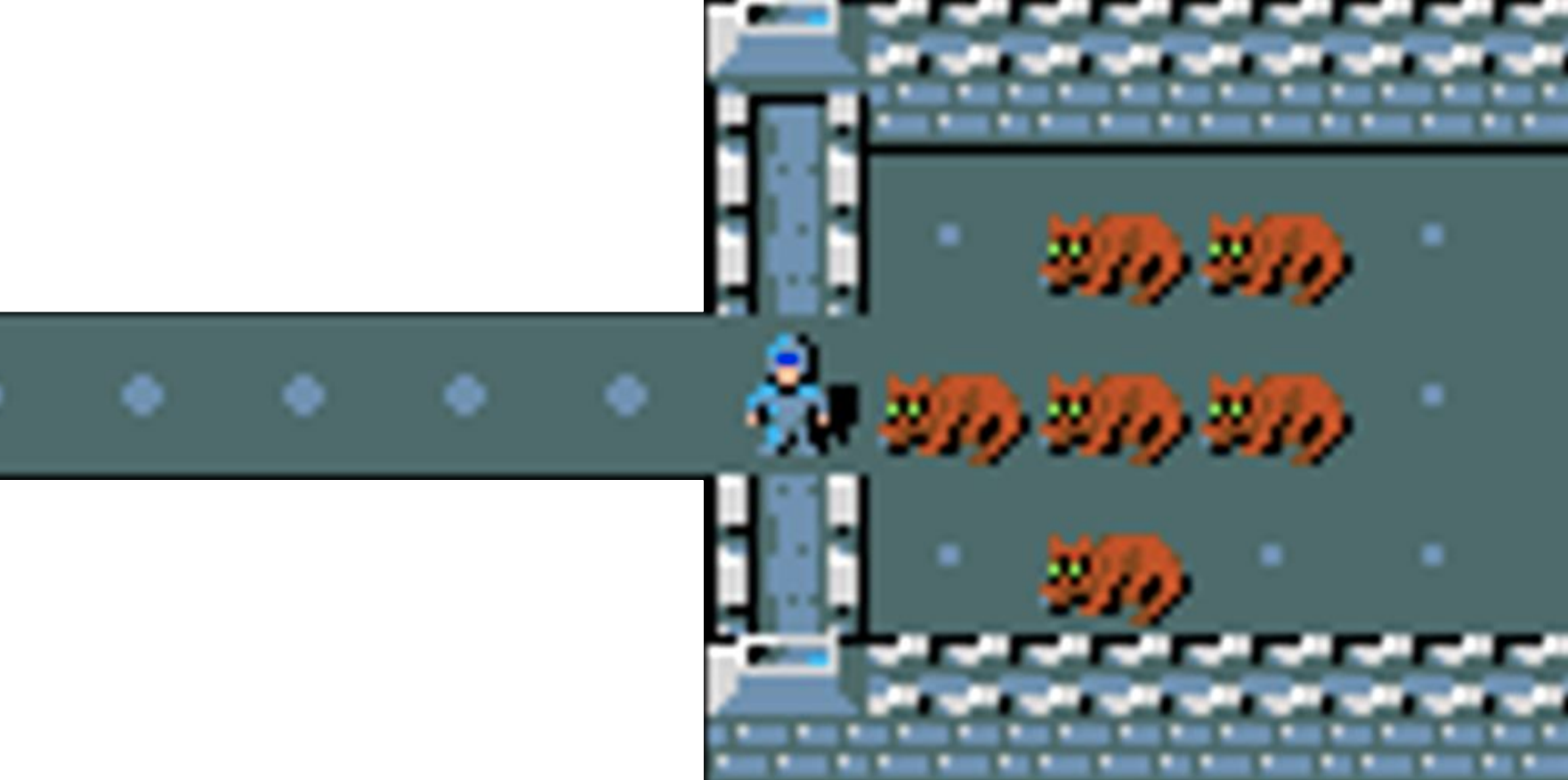}
        \caption{\footnotesize \texttt{CorridorBattle} requires luring monsters into a corridor and fighting them one at a time.}
        \label{fig:corr_fight}
      \end{subfigure}\\
      \begin{subfigure}[b]{0.29\columnwidth}
        \includegraphics[width=.8\textwidth]{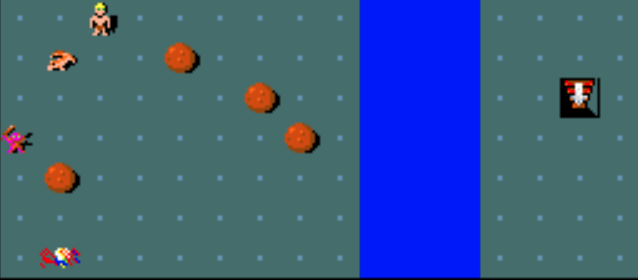}
        \caption{\footnotesize \texttt{River} requires pushing boulders into a river to reach the goal via the generated bridge.}
        \label{fig:river}
      \end{subfigure}
    \end{tabular}
    &
    \begin{subfigure}[b]{0.41\columnwidth}
      \includegraphics[width=\textwidth]{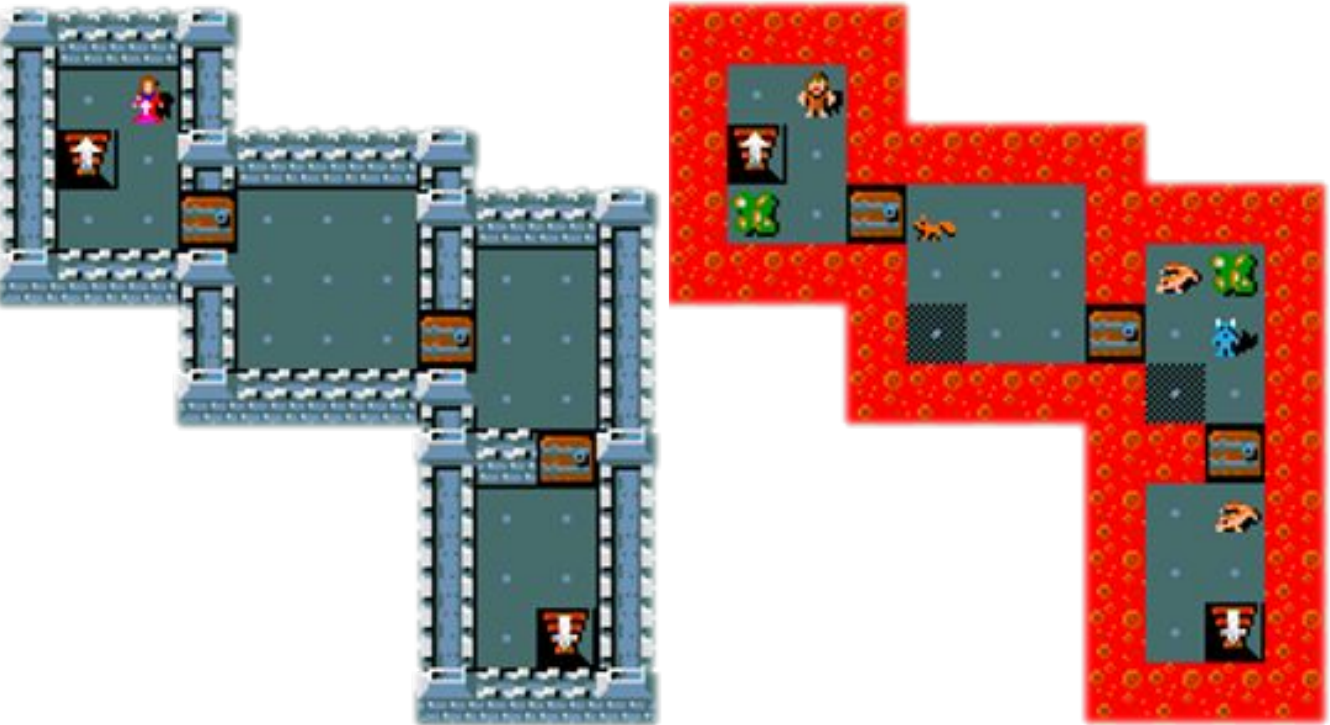}
      \caption{\footnotesize Two versions of \texttt{MultiRoom-N4-S5} task. (left) Regular version (right) Extreme version that includes random monsters, locked doors, and lava tiles instead of walls.}
      \label{fig:multirom}
    \end{subfigure}
    &
    \begin{subfigure}[b]{0.2\columnwidth}
      \includegraphics[width=\textwidth]{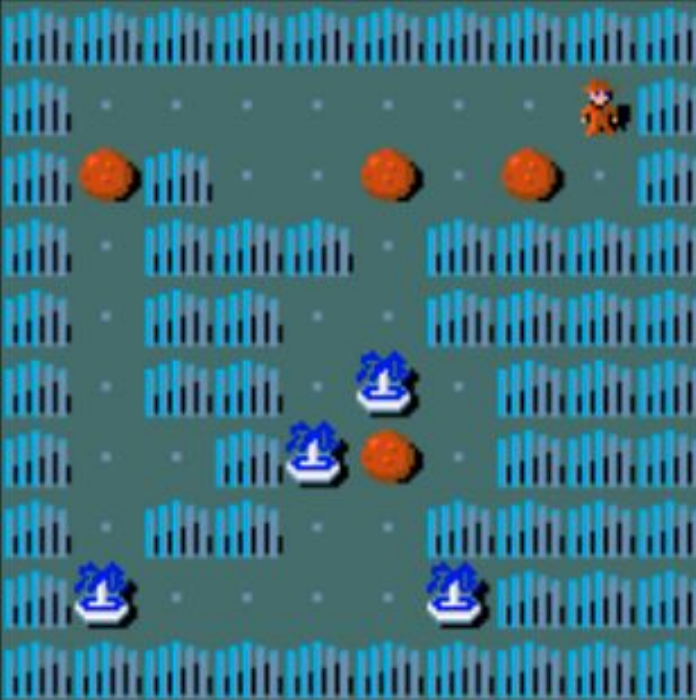}
      \caption{\footnotesize \texttt{Boxoban} requires pushing boulders into different goals (here represented as four fountains).}
      \label{fig:boxoban}
    \end{subfigure}
  \end{tabular}
  \label{fig:screenshots}
  \caption{Screenshots of several MiniHack tasks.}\label{fig:task_images}
    \vspace{-0.5cm}
\end{figure}

\paragraph{Navigation Tasks.}\label{sec:nav_tasks}

MiniHack's navigation tasks challenge the agent to reach the goal position by overcoming various difficulties on their way, such as fighting monsters in corridors (see \cref{fig:corr_fight}), crossing a river by pushing boulders into it (see \cref{fig:river}), navigating through complex or procedurally generated mazes (see \cref{fig:procgen} \textbf{Top} and \textbf{Medium}).
These tasks feature a relatively small action space.\footnote{Movement towards $8$ compass directions, and based on the environment, search, kick, open, and eat actions.} Furthermore, they can be easily extended or adjusted with minimal effort by either changing their definition in Python or the corresponding \desfile{}.
For instance, once the initial version of the task is mastered, one can add different types of monsters, traps or dungeon features, or remove simplifying assumptions (such as having a fixed map layout or full observability), to further challenge RL methods. 
Our suite of 44 diverse environments is meant to assess several of the core capabilities of RL agents, such as exploration, planning, memory, and generalisation.  The detailed descriptions of all navigation tasks, as well as the full list of 44 registered environments, can be found in \cref{appendix:nav_tasks}.

\paragraph{Skill Acquisition Tasks.}\label{sec:skill_tasks}

Our skill acquisition tasks enable utilising the rich diversity of NetHack objects, monsters and dungeon features, and the interactions between them.
These tasks are different from navigation tasks in two ways. First, the skill acquisition tasks feature a large action space (75 actions), where the actions are instantiated differently depending on which object they are acting on. 
Given the large number of entities in MiniHack, usage of objects with an appropriate action in the setting of sparse rewards is extremely difficult, requiring a thorough exploration of the joint state-action space.\footnote{Most of the state-of-the-art exploration methods, such as RND \cite{burda2019exploration}, RIDE \cite{raileanu2020ride}, BeBold \cite{zhang2020bebold}, and AGAC \cite{agac}, rely on state space exploration rather than the state-action space exploration.}
Second, certain actions in skill acquisition tasks are factorised autoregresively~\citep{pierrot2021factored}, i.e., require performing a sequence of follow-up actions for the initial action to have an effect. For example, to put on a ring, the agent needs to select the \texttt{PUTON} action, choose the ring from the inventory and select which hand to put it on. 
As a result, MiniHack allows getting rid of simplifying assumptions that many RL environments impose, such as having a single "schema" action used with a variety of objects regardless of the context.
For the full list of tasks, see \cref{appendix:skill_tasks}.

\paragraph{Porting Existing Environments to MiniHack.}\label{sec:Porting}

Transitioning to using a new environment or benchmark for RL research can be troublesome as it becomes more difficult to compare with prior work that was evaluated on previous environments or benchmarks. Here, we show that prior benchmarks such as MiniGrid \cite{gym_minigrid} and Boxoban \cite{boxobanlevels} can be ported to MiniHack. While the MiniHack versions of these tasks are not visually identical to the originals, they still test the same capabilities as the original versions, which enables researchers familiar with the original tasks to easily analyse the behaviour of agents in these new tasks.
Due to the flexibility and richness of MiniHack, we can incrementally add complexity to the levels in these previous benchmarks and assess the limits of current methods. 
This is especially useful for MiniGrid, where current methods are able to solve all existing tasks \cite{agac, zhang2020bebold}. %

As an example, we present how to patch the navigation tasks in MiniGrid \cite{gym_minigrid} and increase their complexity by adding monsters, locked doors, lava tiles, etc (see \cref{fig:multirom}). 
Similarly, we make use of publicly available levels of Boxoban \cite{boxobanlevels} to offer these task in MiniHack (see \cref{fig:boxoban}). Once ported to MiniHack, these levels can easily be extended, for example, by adding monsters to fight while solving puzzles.
An added benefit of porting such existing benchmarks is that they can use a common observation and action space. This enables investigating transfer learning and easily benchmarking a single algorithm or architecture on a wide variety of challenges, such as the planning problems present in Boxoban and the sparse-reward exploration challenges of MiniGrid.

While MiniHack has the ability to replace a large set of entities ported from original environments, it is worth noting that not all entities have identical replacements. For example, MiniGrid's \texttt{KeyCorridor} includes keys and doors of different colours, whereas the corresponding objects in NetHack have no colour. The randomly moving objects in MiniGrid's \texttt{Dynamic-Obstacles} tasks are also absent.

Despite MiniHack versions of ported environments having minor differences compared to originals, they nonetheless assess the exact same capabilities of RL agents. In particular, while the underlying dynamics of the environment are identical to the original in the case of Boxoban, our MiniGrid version includes slight changes to the agent's action space (turning and moving forwards vs only directional movement) and environment dynamics (the event of opening doors is probabilistic in MiniHack).

\subsection{Evaluation Methodology}\label{appendix:eval_methodology}

Here we describe the evaluation methodology and experimental practice we take in this paper in more detail, as a recommendation for future work evaluating methods on the MiniHack suite of tasks. To ensure a fair comparison between methods, performance should be evaluated in standard conditions. Specifically to MiniHack, this means using the same: action space,\footnote{Larger action spaces can often increase the difficulty of the task.} observation keys,\footnote{Abstractly, the same observation space. There are multiple (multimodal) options which may change the difficulty of the task.} fixed episode length, reward function, game character,\footnote{The default character in MiniHack is a chaotic human male rogue (\texttt{rog-hum-cha-mal}) for navigation tasks and a neutral human male caveman (\texttt{cav-hum-new-mal}) for skill acquisition tasks. For the \texttt{CorridorBattle} tasks, we override the default and use a lawful human female knight (\texttt{kni-hum-law-fem}) instead.} or any other environment parameter that can potentially affect the difficulty of tasks.
When reporting results, we recommend reporting the median reward achieved over at least 5 independent training runs with different seeds. Reporting the median avoids the effects of any outliers, and five independent runs strike a balance between statistical validity and computational requirements. 
When evaluating the generalisation, agents should be trained on a limited number of seeds and tested on held-out seeds or the full distribution of levels.
As well as sharing final performance, sharing full learning curves (provide all independent runs or present an error metric, such as standard deviation), wall-clock time and examples of behaviour are recommended as it can be helpful to other researchers building on or comparing against the results.
If additional information has been used during training or testing, then that should be made clear in any comparisons with other work.
\section{Experiments}\label{sec:Experiments}

In this section, we present experiments on tasks described in \cref{sec:Tasks}. 
The purpose of these experiments is to assess various capabilities of RL agents, and highlight how incremental changes in the environments using the rich entities in MiniHack further challenge existing RL methods. 
We highlight and discuss results on several different tasks, while results for all tasks can be found in \cref{appendix:results}.

Our main baseline for all tasks is based on IMPALA \cite{espeholt2018impala}, as implemented in TorchBeast \cite{torchbeast2019}. For navigation tasks, we also use two popular exploration techniques -- Random Network Distillation \cite[RND,][]{burda2019exploration} and RIDE \cite{raileanu2020ride}, the latter being designed specifically for procedurally generated environments where an agent is unlikely to visit a state more than once.
We train and test agents on the full distribution of levels in each environment.
We make use of the same agent architecture as in \cite{kuttler2020nethack}. %
All details on agent architectures and training setting are described in \cref{appendix:experiments}.
Full results on all MiniHack tasks are available in \cref{appendix:results}.

\begin{figure}
\centering
\includegraphics[width=0.28\textwidth]{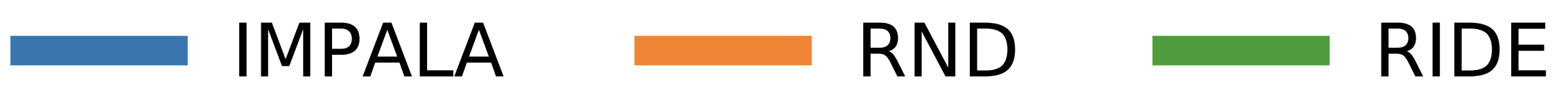}\\
\includegraphics[width=\textwidth]{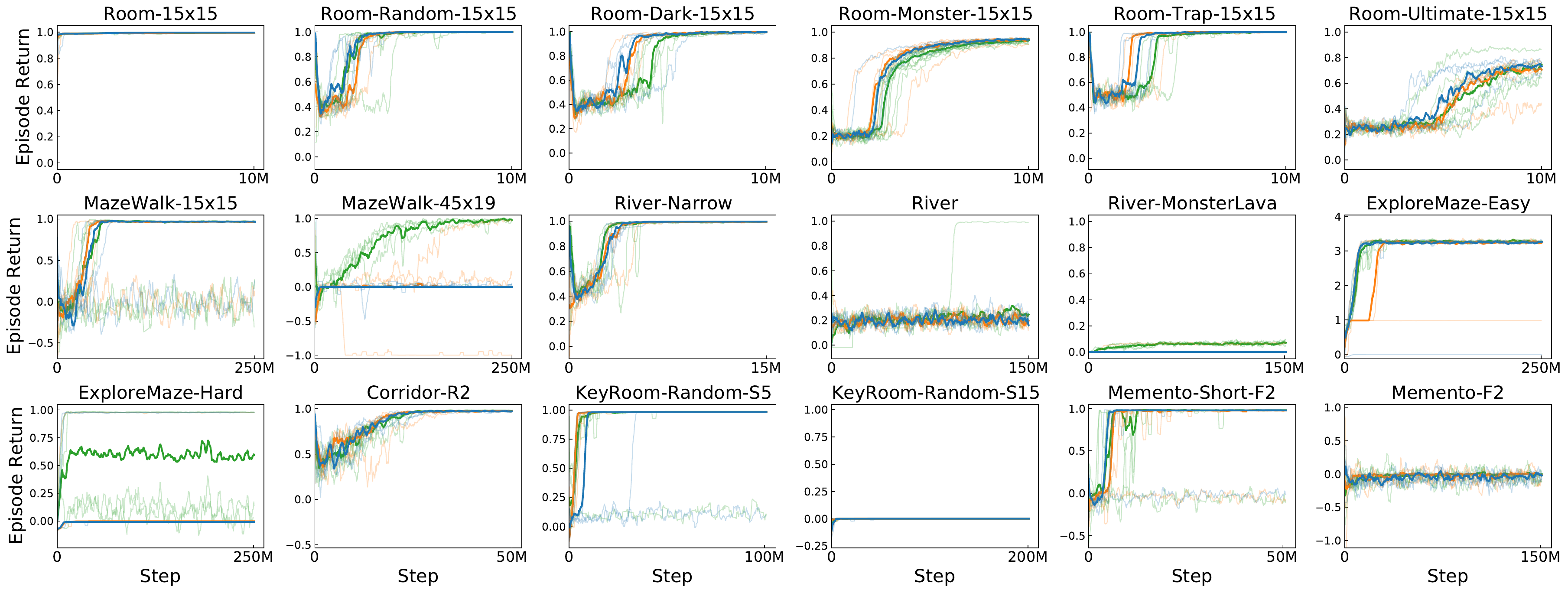}\\
\caption{Mean episode returns on several MiniHack navigation tasks across five independent runs. The median of the runs is bolded.}
\label{fig:nav_results}
\end{figure}

\paragraph{Navigation Tasks.}
\cref{fig:nav_results} summarises results on various challenging MiniHack navigation tasks. While simpler versions of the tasks are often quickly solved by the baseline approaches, adding layers of complexity (such as increasing the size of procedurally generated mazes, resorting to partially observable settings, and adding monsters and traps) renders the baselines incapable of making progress. For example, our baselines fail to get any reward on the most difficult version of the \texttt{River} task that includes moving monsters and deadly lava tiles, challenging the exploration, planning and generalisation capabilities of the agent. 
The results on the \texttt{KeyRoom} tasks highlight the inability of RL methods to handle generalisation at scale. Though the smaller version of the task (\texttt{KeyRoom-Random-S5}) is solved by all baselines, the larger variant (\texttt{KeyRoom-Random-S15}) is not solved by any of the methods.

\begin{figure}
\centering
\includegraphics[width=0.1\textwidth]{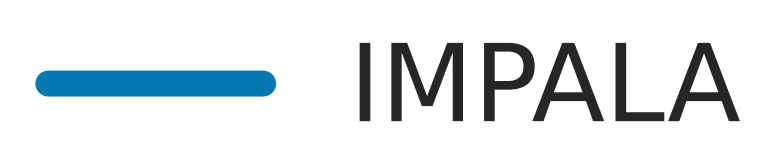}\\
\includegraphics[width=\textwidth]{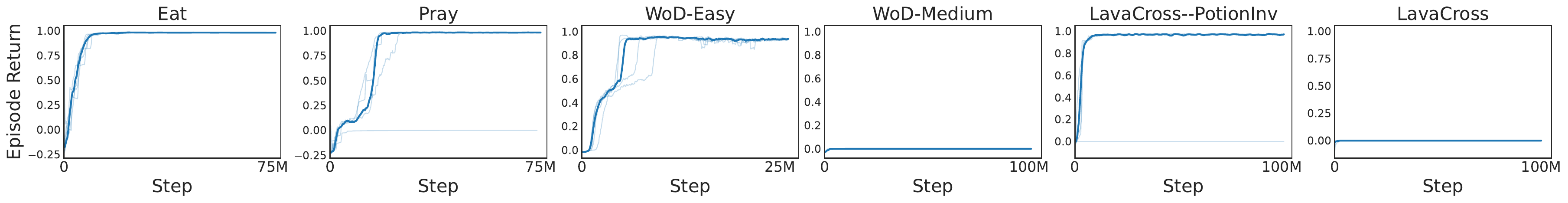}\\
\caption{Mean episode returns on several skill acquisition tasks across five independent runs. The median of the runs is bolded.}
\label{fig:skill_results}
\end{figure}

\paragraph{Skill Acquisition Tasks.}
\cref{fig:skill_results} presents our results on various skill acquisition tasks. 
While the simple tasks that require discovering interaction between a single entity and an action of the agent (e.g., eating comestibles, zapping a wand of death towards a monster, etc.) can be solved by the baselines, the discovery of a sequence of entity-action relations in the presence of environment randomisation and distracting entities remains challenging for RL methods. For instance, none of the runs is able to make progress on \texttt{WoD-Medium} or \texttt{LavaCross} tasks due to insufficient state-action space exploration despite mastering the simplified versions of them.

\begin{figure}
\centering
\includegraphics[width=0.28\textwidth]{figures/results/legend_2.png}\\
\includegraphics[width=.9\textwidth]{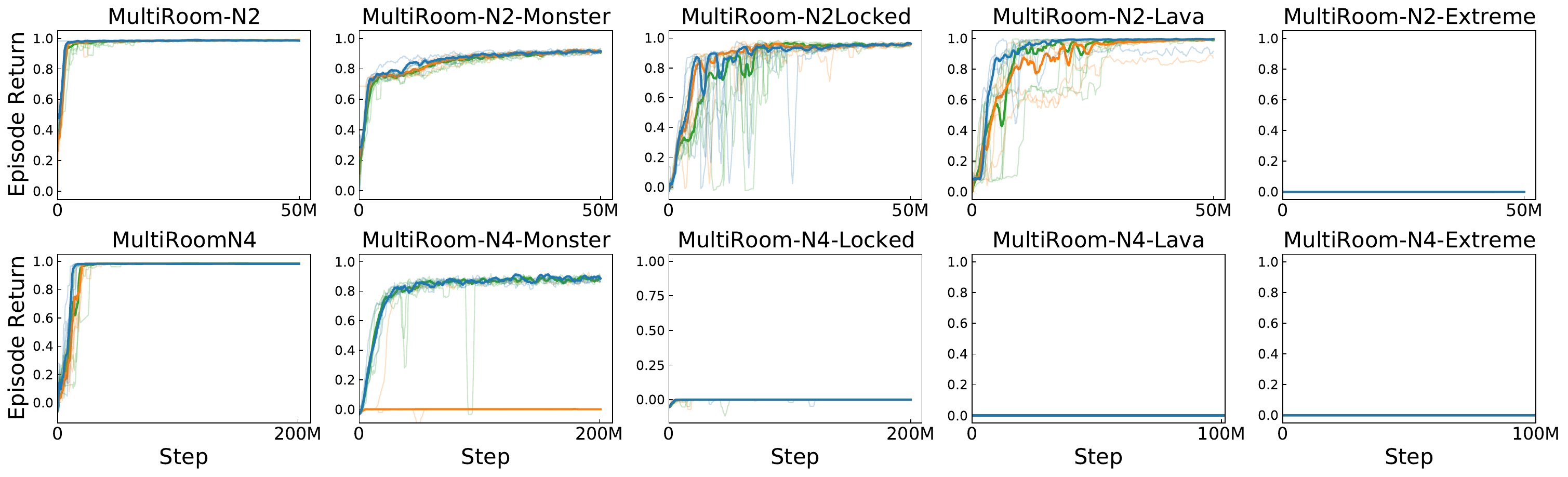}\\
\caption{Mean episode returns on various \texttt{MultiRoom} environments ported from MiniGrid \citep{gym_minigrid} across five independent runs. The median of the runs is bolded.}
\label{fig:multiroom_results}
\vspace{-3mm}
\end{figure}

\paragraph{Ported Environments.} \cref{fig:multiroom_results} presents the results of different versions of the \texttt{MultiRoom} environment ported from MiniGrid \cite{gym_minigrid}. In the version with two rooms, adding additional layers of complexity, such as locked doors, monsters, or lava tiles instead of walls, makes the learning more difficult compared to regular versions of the task. In the Extreme version with all the aforementioned complexities, there is no learning progress at all.
In the version with four rooms, the baseline agents only make progress on the simplest version and the version with added monsters. All additional complexities are beyond the capabilities of baseline methods due to the hard exploration that they require. Consequently, all independent runs fail to obtain any positive reward. These results highlight the ability of MiniHack to assess the limits of RL algorithms in an extendable, well-controlled setting.

\begin{figure}[t!]
\centering
\includegraphics[width=.9\textwidth]{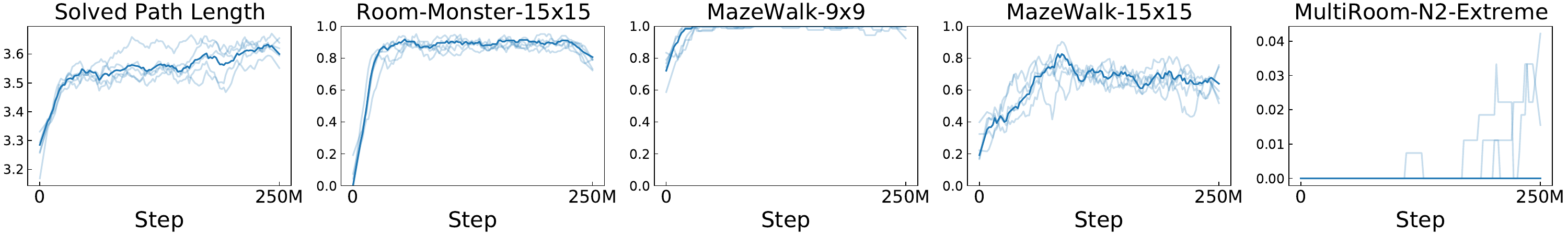}
\caption{Results from the PAIRED algorithm, showing the solved path length of UED environments and zero-shot transfer performance. Plots show five independent runs with the median bolded.}
\label{fig:paired}
\end{figure}

\paragraph{Unsupervised Environment Design.}\label{sec:env_design}

MiniHack also enables research in \emph{Unsupervised Environment Design} (UED), whereby an adaptive task distribution is learned during training by dynamically adjusting free parameters of the task MDP.
MiniHack allows overriding the description file of the environment, making it easy to adjust the MDP configuration as required by UED. 
To test UED in MiniHack, we implement the recent PAIRED algorithm \cite{paired}, which trains an environment adversary to generate environments in order to ultimately train a robust \emph{protagonist} agent, by maximizing the regret, defined as the difference in return between a third, \emph{antagonist} agent and the protagonist. We allow our adversary to place four objects in a small 5x5 room: \{\texttt{walls}, \texttt{lava}, \texttt{monster}, \texttt{locked door}\}. As a result of the curriculum induced by PAIRED, the protagonist is able to improve zero-shot performance on challenging out-of-distribution environments such as \texttt{MultiRoom-N2-Extreme}, despite only ever training on a much smaller environment (see \cref{fig:paired}).

\begin{figure}[t!]
\centering
\includegraphics[width=.9\textwidth]{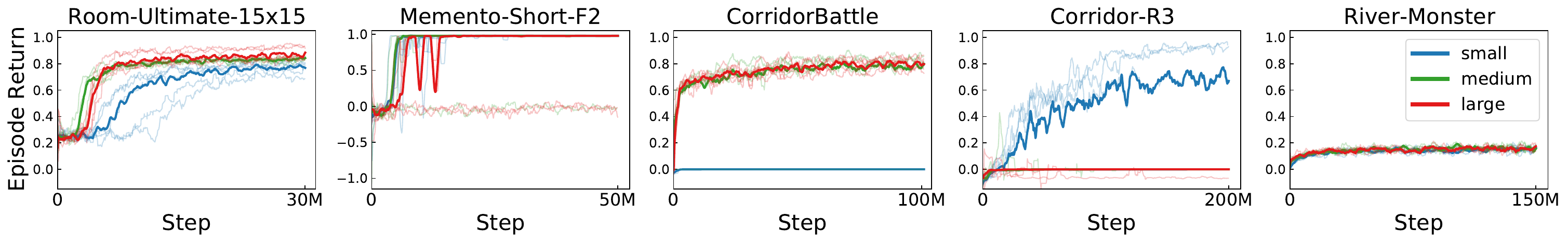}\\
\caption{Mean episode returns of a baseline IMPALA agent using three model architectures.}
\label{fig:arch_comparison}
\vspace{-3mm}
\end{figure}

\paragraph{Agent Architecture Comparison.}We perform additional experiments to compare the performance of the baseline IMPALA agent using different neural architectures. \cref{fig:arch_comparison} presents results using three architectures (small, medium, and large) on selected MiniHack tasks which differ in the number of convolutional layers, the size of hidden MLP layers, as well as the entity embedding dimension (see \cref{appendix:arch_comparison} for full details). The performances of medium and large agent architectures are on par with each other across all five tasks. Interestingly, the small model demonstrates poor performance on \texttt{Room-Ultimate-15} and \texttt{CorridorBattle} environments, but outperforms larger models on the \texttt{Corridor-3} task. While it is known that small models can outperform larger models (both in terms of depth and width) depending on the complexity of the environment \citep{andrychowicz2021what, henderson2017deep}, MiniHack opens door to investigate this phenomenon in a more controlled setting due to the generous environment customisation abilities it provides. We thus believe MiniHack would be a useful testbed for exploring the impact of architecture on RL agent training.

\begin{wrapfigure}{r}{0.3\textwidth}
\vspace{-8mm}
\includegraphics[width=\linewidth]{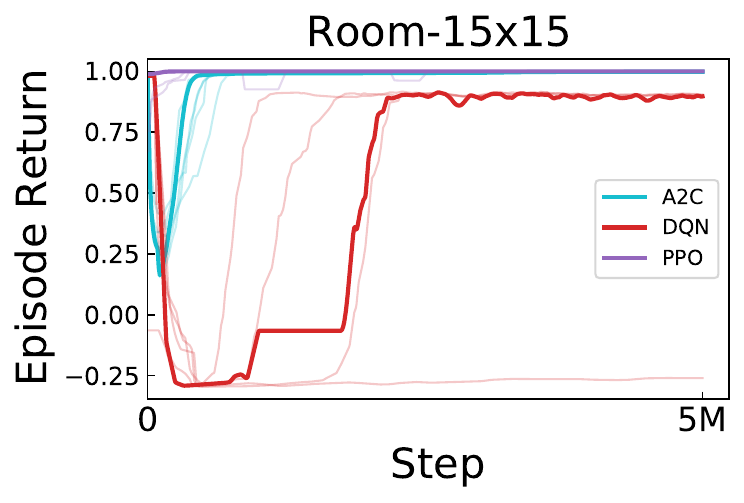}
\vspace{-6mm}
\caption{RLlib results.}
\label{fig:rllib}
\vspace{-7mm}
\end{wrapfigure}

\paragraph{RLlib Example.} To help kickstart the development of RL models using MiniHack, we also provide integration with RLlib~\cite{pmlr-v80-liang18b}. RLlib enables using a wide range of RL algorithms within the framework, ensuring that research on MiniHack can be performed with varying computational constraints. \cref{fig:rllib} presents the results of DQN~\cite{mnih2015human}, A2C~\cite{mnih2016asynchronous}, and PPO~\cite{Schulman2017ProximalPO} methods on the \texttt{Room-15x15} task. See \cref{appendix:rllib} for more details.

\vspace{-2mm}
\section{Related Work}\label{sec:related_work}
\vspace{-1mm}

The RL community has made extensive use of video games as testbeds for RL algorithms \citep{ale, vinyals2017starcraft, openai2019dota, nichol2018gotta, kempka2016vizdoom}. 
However, such games are computationally expensive to run and not easily modifiable. Furthermore, they provide only a fixed set of levels, which results in the overfitting of trained policies.
ProcGen~\cite{cobbe2019procgen} partially addresses this issue by providing a collection of 2D games with procedurally generated levels. %
However, the richness of games is still limited and only minor modifications are possible, unlike the rich environment creation capabilities that MiniHack provides. Based on the C/C++ engines of NetHack and \NLE{}, MiniHack is 16 times faster than ALE~\cite{ale} and faster than ProcGen by approximately 10\% (see Appendix D of \cite{kuttler2020nethack} for an approximate comparison). %

MiniGrid \cite{gym_minigrid} addresses the issue of computational efficiency by providing a collection of procedurally generated grid-world tasks. Nevertheless, the complexity of the environments is still limited, containing only a few types of entities and small action space. Moreover, extending the existing environments is difficult as it requires understanding the internals of the library. %
MiniHack provides a much richer set of entities (hundreds of objects, monsters, dungeon features) and a much larger action space. Moreover, MiniHack is designed to be easy to extend and build on top of, only requiring familiarity with a flexible and expressive high-level DSL but no underlying implementation details.

\textit{bsuite}~\citep{osband2020bsuite} features a set of simple environments designed to test specific capabilities of RL agents, such as memory or exploration.
In contrast, MiniHack is not confined to a static task set allowing researchers to easily extend the existing task set without the need to understand the implementation details of the framework.

Several roguelike games, a genre of video games characterised by progressing through procedurally generated dungeon levels and grid-based movements, have been proposed as RL benchmarks. 
Rogueinabox~\cite{asperti2017rogueinabox} provides an interface to Rogue, a roguelike game with simple dynamics and limited complexity. 
Rogue-Gym~\cite{kanagawa2019rogue} introduces a simple roguelike game built for evaluating generalisation in RL and uses parameterisable generation of game levels.
\NLE{} \cite{kuttler2020nethack} and \textit{gym-nethack}~\cite{campbell-17-learning, campbell2018exploration} provide a Gym interface around the game of NetHack. However, these benchmarks use either fixed, predefined level descriptions of the full games, or a fixed set of concrete subtask (e.g., 1-on-1 combat with individual monsters \cite{campbell2018exploration}).
In contrast, MiniHack allows easily customising dungeon layouts and placement of environment features, monsters and objects by a convenient Python interface or a human-readable description language. 

MiniHack is not the first to provide a sandbox for developing environments.
PyVGDL \cite{schaul_video_2013, schaul_extensible_2014} uses a concrete realisation of the Video Game Description Language \cite[VGDL,][]{ebner_towards_2013} for creating 2D video games. %
The original software library of PyVGDL is no longer supported, while the 2.0 version is under development~\cite{pyvgdl2}. 
The GVGAI framework~\citep{GVGAI} is also based on the VGDL but suffers from a computational overhead due to its Java implementation.
Griddly~\citep{griddly} provides a highly configurable mechanism for designing diverse environments using a custom description language. 
Similar to MiniHack, Griddly is based on an efficient C/C++ core engine and is fast to run experiments on. Griddly is approximately an order of magnitude faster than MiniHack for simple environments, but it is unclear to what extent adding complex dynamics to Griddly, equivalent to what MiniHack provides, will decrease its speed. 
Furthermore, Griddly supports multi-agent and real-time strategy (RTS) games, unlike MiniHack.
While PyVGDL, GVGAI, and Griddly can be used to create various entities, developing rich environment dynamics requires a significant amount of work. In contrast, MiniHack features a large collection of predefined objects, items, monsters, environment features, spells, etc and complex environment mechanics from the game of NetHack, thus hitting the sweet spot between customizability and the richness of entities and environment dynamics to draw upon. %
MiniHack also provides a rich set of multimodal observations (textual descriptions, symbolic and pixel-based observations) and a convenient Python interface for describing environments in only a few lines of code. 
Finally, the Malmo Platform~\citep{DBLP:conf/ijcai/JohnsonHHB16} and MineRL~\citep{guss2019minerlcomp} provides an interface to a popular game of Minecraft. 
While being rich in the environment dynamics, Minecraft is computationally intensive compared to NetHack~\citep{raymond2020guide} (MiniHack is approximately 240 times faster than MineRL~\cite{kuttler2020nethack}).

\vspace{-3mm}
\section{Conclusion}
\vspace{-2mm}
In this work, we presented MiniHack, an easy-to-use framework for creating rich and varied RL environments, as well as a suite of tasks developed using this framework. Built upon NLE and the \desfile{} format, MiniHack enables the use of rich entities and dynamics from the game of NetHack to create a large variety of RL environments for targeted experimentation, while also allowing painless scaling-up of the difficulty of existing environments. MiniHack's environments are procedurally generated by default, ensuring the evaluation of systematic generalization of RL agents.
The suite of tasks we release with MiniHack tests the limits of RL methods and enables researchers to test a wide variety of capabilities in a unified experimental setting.
To enable further experimentation and research, we open-source our code for training agents with numerous RL algorithms, outline best practices on how to evaluate algorithms on the benchmark suite of tasks, and provide baseline results on all of the tasks that we released.

\section*{Acknowledgements}

We thank Danielle Rothermel, Zhengyao Jiang, Pasquale Minervini, Vegard Mella, Olivier Teytaud, and Luis Pineda for insightful discussions and valuable feedback on this work. We also thank our anonymous reviewers for their recommendations on improving this paper.

\bibliographystyle{plain}
\bibliography{bib/minihack.bib}

\begin{thebibliography}{10}

\bibitem{andrychowicz2021what}
Marcin Andrychowicz, Anton Raichuk, Piotr Sta{\'n}czyk, Manu Orsini, Sertan
  Girgin, Rapha{\"e}l Marinier, Leonard Hussenot, Matthieu Geist, Olivier
  Pietquin, Marcin Michalski, Sylvain Gelly, and Olivier Bachem.
\newblock What matters for on-policy deep actor-critic methods? a large-scale
  study.
\newblock In {\em International Conference on Learning Representations}, 2021.

\bibitem{asperti2017rogueinabox}
Andrea Asperti, Carlo De~Pieri, and Gianmaria Pedrini.
\newblock Rogueinabox: an environment for roguelike learning.
\newblock {\em International Journal of Computers}, 2, 2017.

\bibitem{Agent57}
Adri{\`{a}}~Puigdom{\`{e}}nech Badia, Bilal Piot, Steven Kapturowski, Pablo
  Sprechmann, Alex Vitvitskyi, Daniel Guo, and Charles Blundell.
\newblock Agent57: Outperforming the atari human benchmark.
\newblock In {\em Proceedings of the 37th International Conference on Machine
  Learning}, 2020.

\bibitem{griddly}
Chris Bamford, Shengyi Huang, and Simon~M. Lucas.
\newblock Griddly: {A} platform for {AI} research in games.
\newblock {\em CoRR}, abs/2011.06363, 2020.

\bibitem{BeattieLTWWKLGV16}
Charles Beattie, Joel~Z. Leibo, Denis Teplyashin, Tom Ward, Marcus Wainwright,
  Heinrich K{\"{u}}ttler, Andrew Lefrancq, Simon Green, V{\'{\i}}ctor
  Vald{\'{e}}s, Amir Sadik, Julian Schrittwieser, Keith Anderson, Sarah York,
  Max Cant, Adam Cain, Adrian Bolton, Stephen Gaffney, Helen King, Demis
  Hassabis, Shane Legg, and Stig Petersen.
\newblock Deepmind lab.
\newblock {\em CoRR}, abs/1612.03801, 2016.

\bibitem{bellemare2016unifying}
Marc Bellemare, Sriram Srinivasan, Georg Ostrovski, Tom Schaul, David Saxton,
  and Remi Munos.
\newblock Unifying count-based exploration and intrinsic motivation.
\newblock In {\em NeurIPS}, 2016.

\bibitem{ale}
Marc~G. Bellemare, Yavar Naddaf, Joel Veness, and Michael Bowling.
\newblock The {A}rcade {L}earning {E}nvironment: {A}n {E}valuation {P}latform
  for {G}eneral {A}gents.
\newblock {\em CoRR}, abs/1207.4708, 2012.

\bibitem{DBLP:journals/corr/BrockmanCPSSTZ16}
Greg Brockman, Vicki Cheung, Ludwig Pettersson, Jonas Schneider, John Schulman,
  Jie Tang, and Wojciech Zaremba.
\newblock {OpenAI Gym}.
\newblock {\em CoRR}, abs/1606.01540, 2016.

\bibitem{burda2019exploration}
Yuri Burda, Harrison Edwards, Amos Storkey, and Oleg Klimov.
\newblock Exploration by random network distillation.
\newblock In {\em ICML}, 2019.

\bibitem{campbell-17-learning}
Jonathan Campbell and Clark Verbrugge.
\newblock Learning combat in {N}et{H}ack.
\newblock In {\em AIIDE}, 2017.

\bibitem{campbell2018exploration}
Jonathan Campbell and Clark Verbrugge.
\newblock Exploration in {N}et{H}ack with secret discovery.
\newblock {\em IEEE Transactions on Games}, 2018.

\bibitem{gym_minigrid}
Maxime Chevalier-Boisvert, Lucas Willems, and Suman Pal.
\newblock {Minimalistic Gridworld Environment for OpenAI Gym}.
\newblock \url{https://github.com/maximecb/gym-minigrid}, 2018.

\bibitem{cobbe2019procgen}
Karl Cobbe, Christopher Hesse, Jacob Hilton, and John Schulman.
\newblock Leveraging procedural generation to benchmark reinforcement learning.
\newblock {\em arXiv preprint arXiv:1912.01588}, 2019.

\bibitem{paired}
Michael Dennis, Natasha Jaques, Eugene Vinitsky, Alexandre Bayen, Stuart
  Russell, Andrew Critch, and Sergey Levine.
\newblock Emergent complexity and zero-shot transfer via unsupervised
  environment design.
\newblock In {\em Advances in Neural Information Processing Systems},
  volume~33, 2020.

\bibitem{dennis2020emergent}
Michael Dennis, Natasha Jaques, Eugene Vinitsky, Alexandre Bayen, Stuart
  Russell, Andrew Critch, and Sergey Levine.
\newblock Emergent complexity and zero-shot transfer via unsupervised
  environment design.
\newblock In H.~Larochelle, M.~Ranzato, R.~Hadsell, M.~F. Balcan, and H.~Lin,
  editors, {\em Advances in Neural Information Processing Systems}, volume~33,
  pages 13049--13061. Curran Associates, Inc., 2020.

\bibitem{dulac_arnold2020empirical}
Gabriel Dulac{-}Arnold, Nir Levine, Daniel~J. Mankowitz, Jerry Li, Cosmin
  Paduraru, Sven Gowal, and Todd Hester.
\newblock An empirical investigation of the challenges of real-world
  reinforcement learning.
\newblock {\em CoRR}, abs/2003.11881, 2020.

\bibitem{ebner_towards_2013}
Marc Ebner, John Levine, Simon~M. Lucas, Tom Schaul, Tommy Thompson, and Julian
  Togelius.
\newblock Towards a {Video} {Game} {Description} {Language}.
\newblock page 16 pages, 2013.
\newblock Artwork Size: 16 pages Medium: application/pdf Publisher: Schloss
  Dagstuhl - Leibniz-Zentrum fuer Informatik GmbH, Wadern/Saarbruecken, Germany
  Version Number: 1.0.

\bibitem{ecoffet2019go}
Adrien Ecoffet, Joost Huizinga, Joel Lehman, Kenneth~O. Stanley, and Jeff
  Clune.
\newblock {Go-Explore: A New Approach for Hard-exploration Problems}.
\newblock {\em arXiv preprint arXiv:1901.10995}, 2019.

\bibitem{espeholt2018impala}
Lasse Espeholt, Hubert Soyer, Remi Munos, Karen Simonyan, Vlad Mnih, Tom Ward,
  Yotam Doron, Vlad Firoiu, Tim Harley, Iain Dunning, et~al.
\newblock Impala: Scalable distributed deep-rl with importance weighted
  actor-learner architectures.
\newblock In {\em International Conference on Machine Learning}, pages
  1407--1416. PMLR, 2018.

\bibitem{agac}
Yannis Flet{-}Berliac, Johan Ferret, Olivier Pietquin, Philippe Preux, and
  Matthieu Geist.
\newblock Adversarially guided actor-critic.
\newblock {\em CoRR}, abs/2102.04376, 2021.

\bibitem{boxobanlevels}
Arthur Guez, Mehdi Mirza, Karol Gregor, Rishabh Kabra, Sebastien Racaniere,
  Theophane Weber, David Raposo, Adam Santoro, Laurent Orseau, Tom Eccles, Greg
  Wayne, David Silver, Timothy Lillicrap, and Victor Valdes.
\newblock An investigation of model-free planning: boxoban levels.
\newblock https://github.com/deepmind/boxoban-levels/, 2018.

\bibitem{guo2020memory}
Yijie Guo, Jongwook Choi, Marcin Moczulski, Shengyu Feng, Samy Bengio, Mohammad
  Norouzi, and Honglak Lee.
\newblock Memory based trajectory-conditioned policies for learning from sparse
  rewards.
\newblock {\em Advances in Neural Information Processing Systems}, 33, 2020.

\bibitem{guss2019minerlcomp}
William~H. Guss, Cayden Codel, Katja Hofmann, Brandon Houghton, Noboru Kuno,
  Stephanie Milani, Sharada Mohanty, Diego~Perez Liebana, Ruslan Salakhutdinov,
  Nicholay Topin, et~al.
\newblock The {M}ine{RL} competition on sample efficient reinforcement learning
  using human priors.
\newblock {\em NeurIPS Competition Track}, 2019.

\bibitem{henderson2017deep}
Peter Henderson, Riashat Islam, Philip Bachman, Joelle Pineau, Doina Precup,
  and David Meger.
\newblock Deep reinforcement learning that matters.
\newblock {\em CoRR}, abs/1709.06560, 2017.

\bibitem{Hill2020Environmental}
Felix Hill, Andrew Lampinen, Rosalia Schneider, Stephen Clark, Matthew
  Botvinick, James~L. McClelland, and Adam Santoro.
\newblock Environmental drivers of systematicity and generalization in a
  situated agent.
\newblock In {\em International Conference on Learning Representations}, 2020.

\bibitem{DBLP:journals/neco/HochreiterS97}
Sepp Hochreiter and J{\"{u}}rgen Schmidhuber.
\newblock Long short-term memory.
\newblock {\em Neural Computation}, 9(8):1735--1780, 1997.

\bibitem{ibarz2021lessons}
Julian Ibarz, Jie Tan, Chelsea Finn, Mrinal Kalakrishnan, Peter Pastor, and
  Sergey Levine.
\newblock How to train your robot with deep reinforcement learning: lessons we
  have learned.
\newblock {\em The International Journal of Robotics Research},
  40(4-5):698--721, 2021.

\bibitem{DBLP:conf/ijcai/JohnsonHHB16}
Matthew Johnson, Katja Hofmann, Tim Hutton, and David Bignell.
\newblock The malmo platform for artificial intelligence experimentation.
\newblock In {\em IJCAI}, 2016.

\bibitem{juliani2019obstacle}
Arthur Juliani, Ahmed Khalifa, Vincent{-}Pierre Berges, Jonathan Harper, Ervin
  Teng, Hunter Henry, Adam Crespi, Julian Togelius, and Danny Lange.
\newblock Obstacle tower: {A} generalization challenge in vision, control, and
  planning.
\newblock In Sarit Kraus, editor, {\em Proceedings of the Twenty-Eighth
  International Joint Conference on Artificial Intelligence, {IJCAI} 2019,
  Macao, China, August 10-16, 2019}, pages 2684--2691. ijcai.org, 2019.

\bibitem{kanagawa2019rogue}
Yuji Kanagawa and Tomoyuki Kaneko.
\newblock Rogue-gym: A new challenge for generalization in reinforcement
  learning.
\newblock In {\em IEEE Conference on Games}, 2019.

\bibitem{kempka2016vizdoom}
Micha{\l} Kempka, Marek Wydmuch, Grzegorz Runc, Jakub Toczek, and Wojciech
  Ja{\'s}kowski.
\newblock Vizdoom: A doom-based ai research platform for visual reinforcement
  learning.
\newblock In {\em IEEE Conference on Computational Intelligence and Games},
  2016.

\bibitem{torchbeast2019}
Heinrich K\"{u}ttler, Nantas Nardelli, Thibaut Lavril, Marco Selvatici,
  Viswanath Sivakumar, Tim Rockt\"{a}schel, and Edward Grefenstette.
\newblock {TorchBeast: A PyTorch Platform for Distributed RL}.
\newblock {\em arXiv preprint arXiv:1910.03552}, 2019.

\bibitem{kuttler2020nethack}
Heinrich K{\"u}ttler, Nantas Nardelli, Alexander~H Miller, Roberta Raileanu,
  Marco Selvatici, Edward Grefenstette, and Tim Rockt{\"a}schel.
\newblock The nethack learning environment.
\newblock {\em NeurIPS 2020}, 2020.

\bibitem{pmlr-v80-liang18b}
Eric Liang, Richard Liaw, Robert Nishihara, Philipp Moritz, Roy Fox, Ken
  Goldberg, Joseph Gonzalez, Michael Jordan, and Ion Stoica.
\newblock {RL}lib: Abstractions for distributed reinforcement learning.
\newblock In {\em Proceedings of the 35th International Conference on Machine
  Learning}, volume~80 of {\em Proceedings of Machine Learning Research}, pages
  3053--3062. PMLR, 10--15 Jul 2018.

\bibitem{mnih2016asynchronous}
Volodymyr Mnih, Adria~Puigdomenech Badia, Mehdi Mirza, Alex Graves, Timothy
  Lillicrap, Tim Harley, David Silver, and Koray Kavukcuoglu.
\newblock Asynchronous methods for deep reinforcement learning.
\newblock In {\em International conference on machine learning}, pages
  1928--1937, 2016.

\bibitem{mnih2015human}
Volodymyr Mnih, Koray Kavukcuoglu, David Silver, Andrei~A Rusu, Joel Veness,
  Marc~G Bellemare, Alex Graves, Martin Riedmiller, Andreas~K Fidjeland, Georg
  Ostrovski, et~al.
\newblock Human-level control through deep reinforcement learning.
\newblock {\em nature}, 518(7540):529--533, 2015.

\bibitem{moritz2018ray}
Philipp Moritz, Robert Nishihara, Stephanie Wang, Alexey Tumanov, Richard Liaw,
  Eric Liang, Melih Elibol, Zongheng Yang, William Paul, Michael~I Jordan,
  et~al.
\newblock Ray: A distributed framework for emerging $\{$AI$\}$ applications.
\newblock In {\em 13th $\{$USENIX$\}$ Symposium on Operating Systems Design and
  Implementation ($\{$OSDI$\}$ 18)}, pages 561--577, 2018.

\bibitem{nhwiki}
{NetHack Wiki}.
\newblock {NetHackWiki}.
\newblock \url{https://nethackwiki.com/}, 2020.
\newblock Accessed: 2021-05-01.

\bibitem{des-file}
{NetHack Wiki}.
\newblock {des-file format}.
\newblock \url{https://nethackwiki.com/wiki/Des-file_format}, 2021.
\newblock Accessed: 2021-05-20.

\bibitem{nichol2018gotta}
Alex Nichol, Vicki Pfau, Christopher Hesse, Oleg Klimov, and John Schulman.
\newblock Gotta learn fast: A new benchmark for generalization in rl.
\newblock {\em arXiv preprint arXiv:1804.03720}, 2018.

\bibitem{openai2019dota}
OpenAI, Christopher Berner, Greg Brockman, Brooke Chan, Vicki Cheung,
  Przemysław Dębiak, Christy Dennison, David Farhi, Quirin Fischer, Shariq
  Hashme, Chris Hesse, Rafal Józefowicz, Scott Gray, Catherine Olsson, Jakub
  Pachocki, Michael Petrov, Henrique~Pondé de~Oliveira~Pinto, Jonathan Raiman,
  Tim Salimans, Jeremy Schlatter, Jonas Schneider, Szymon Sidor, Ilya
  Sutskever, Jie Tang, Filip Wolski, and Susan Zhang.
\newblock Dota 2 with large scale deep reinforcement learning.
\newblock {\em arXiv}, abs/1912.06680, 2019.

\bibitem{osband2016deep}
Ian Osband, Charles Blundell, Alexander Pritzel, and Benjamin Van~Roy.
\newblock Deep exploration via bootstrapped dqn.
\newblock In {\em Advances in neural information processing systems}, pages
  4026--4034, 2016.

\bibitem{osband2020bsuite}
Ian Osband, Yotam Doron, Matteo Hessel, John Aslanides, Eren Sezener, Andre
  Saraiva, Katrina McKinney, Tor Lattimore, Csaba {Sz}epesv{\'a}ri, Satinder
  Singh, Benjamin Van~Roy, Richard Sutton, David Silver, and Hado van Hasselt.
\newblock Behaviour suite for reinforcement learning.
\newblock In {\em International Conference on Learning Representations}, 2020.

\bibitem{GVGAI}
Diego Perez-Liebana, Jialin Liu, Ahmed Khalifa, Raluca~D. Gaina, Julian
  Togelius, and Simon~M. Lucas.
\newblock General video game ai: A multitrack framework for evaluating agents,
  games, and content generation algorithms.
\newblock {\em IEEE Transactions on Games}, 11(3):195--214, 2019.

\bibitem{pierrot2021factored}
Thomas PIERROT, Valentin Mac{\'e}, Jean-Baptiste Sevestre, Louis Monier,
  Alexandre Laterre, Nicolas Perrin, Karim Beguir, and Olivier Sigaud.
\newblock Factored action spaces in deep reinforcement learning, 2021.

\bibitem{raileanu2020ride}
Roberta Raileanu and Tim Rockt{\"a}schel.
\newblock {RIDE}: Rewarding impact-driven exploration for
  procedurally-generated environments.
\newblock In {\em ICLR}, 2020.

\bibitem{raymond2020guide}
Eric~S. Raymond, Mike Stephenson, et~al.
\newblock {\em {A Guide to the Mazes of Menace: Guidebook for NetHack}}.
\newblock NetHack DevTeam, 2020.

\bibitem{schaul_video_2013}
Tom Schaul.
\newblock A video game description language for model-based or interactive
  learning.
\newblock In {\em 2013 IEEE Conference on Computational Inteligence in Games
  (CIG)}, pages 1--8, 2013.

\bibitem{schaul_extensible_2014}
Tom Schaul.
\newblock An {Extensible} {Description} {Language} for {Video} {Games}.
\newblock {\em IEEE Transactions on Computational Intelligence and AI in
  Games}, 6(4):325--331, December 2014.
\newblock Conference Name: IEEE Transactions on Computational Intelligence and
  AI in Games.

\bibitem{schaul2015prioritized}
Tom Schaul, John Quan, Ioannis Antonoglou, and David Silver.
\newblock Prioritized experience replay.
\newblock {\em arXiv preprint arXiv:1511.05952}, 2015.

\bibitem{Schulman2017ProximalPO}
John Schulman, Filip Wolski, Prafulla Dhariwal, Alec Radford, and Oleg Klimov.
\newblock Proximal policy optimization algorithms.
\newblock {\em ArXiv}, abs/1707.06347, 2017.

\bibitem{suttonbarto}
Richard~S Sutton, Andrew~G Barto, et~al.
\newblock {\em Introduction to reinforcement learning}, volume 135.
\newblock MIT press Cambridge, 1998.

\bibitem{15hardestVG}
{The Telepraph}.
\newblock {The 25 hardest video games ever}.
\newblock
  \url{https://www.telegraph.co.uk/gaming/what-to-play/the-15-hardest-video-games-ever/nethack/},
  2021.
\newblock Accessed: 2021-05-05.

\bibitem{todorov2012mujoco}
Emanuel Todorov, Tom Erez, and Yuval Tassa.
\newblock Mujoco: A physics engine for model-based control.
\newblock In {\em Intelligent Robots and Systems (IROS), 2012 IEEE/RSJ
  International Conference on}, pages 5026--5033. IEEE, 2012.

\bibitem{van2016deep}
Hado Van~Hasselt, Arthur Guez, and David Silver.
\newblock Deep reinforcement learning with double q-learning.
\newblock In {\em Proceedings of the AAAI Conference on Artificial
  Intelligence}, volume~30, 2016.

\bibitem{pyvgdl2}
Ruben Vereecken.
\newblock {PyVGDL 2.0: A video game description language for AI research}.
\newblock \url{https://github.com/rubenvereecken/py-vgdl}, 2020.

\bibitem{vinyals2017starcraft}
Oriol Vinyals, Timo Ewalds, Sergey Bartunov, Petko Georgiev, Alexander~Sasha
  Vezhnevets, Michelle Yeo, Alireza Makhzani, Heinrich K{\"u}ttler, John
  Agapiou, Julian Schrittwieser, et~al.
\newblock {StarCraft II: {A} New Challenge for Reinforcement Learning}.
\newblock {\em arXiv preprint arXiv:1708.04782}, 2017.

\bibitem{wang2021alchemy}
Jane Wang, Michael King, Nicolas Porcel, Zeb Kurth-Nelson, Tina Zhu, Charlie
  Deck, Peter Choy, Mary Cassin, Malcolm Reynolds, Francis Song, Gavin
  Buttimore, David Reichert, Neil Rabinowitz, Loic Matthey, Demis Hassabis,
  Alex Lerchner, and Matthew Botvinick.
\newblock Alchemy: A structured task distribution for meta-reinforcement
  learning.
\newblock {\em arXiv preprint arXiv:2102.02926}, 2021.

\bibitem{wang2016dueling}
Ziyu Wang, Tom Schaul, Matteo Hessel, Hado Hasselt, Marc Lanctot, and Nando
  Freitas.
\newblock Dueling network architectures for deep reinforcement learning.
\newblock In {\em International conference on machine learning}, pages
  1995--2003. PMLR, 2016.

\bibitem{yu2019meta}
Tianhe Yu, Deirdre Quillen, Zhanpeng He, Ryan Julian, Karol Hausman, Chelsea
  Finn, and Sergey Levine.
\newblock Meta-world: A benchmark and evaluation for multi-task and meta
  reinforcement learning.
\newblock In {\em Conference on Robot Learning (CoRL)}, 2019.

\bibitem{pbt_mujoco}
Baohe Zhang, Raghu Rajan, Luis Pineda, Nathan Lambert, Andr{\'e} Biedenkapp,
  Kurtland Chua, Frank Hutter, and Roberto Calandra.
\newblock On the importance of hyperparameter optimization for model-based
  reinforcement learning.
\newblock In {\em Proceedings of The 24th International Conference on
  Artificial Intelligence and Statistics}, volume 130 of {\em Proceedings of
  Machine Learning Research}. PMLR, 13--15 Apr 2021.

\bibitem{zhang2020bebold}
Tianjun Zhang, Huazhe Xu, Xiaolong Wang, Yi~Wu, Kurt Keutzer, Joseph~E.
  Gonzalez, and Yuandong Tian.
\newblock Bebold: Exploration beyond the boundary of explored regions.
\newblock {\em CoRR}, abs/2012.08621, 2020.

\bibitem{zhangZL15}
Xiang Zhang, Junbo~Jake Zhao, and Yann LeCun.
\newblock Character-level convolutional networks for text classification.
\newblock {\em CoRR}, abs/1509.01626, 2015.

\end{thebibliography}
\newpage

\newpage
\appendix

\section{The \desfile{} Format}\label{appendix:des_file}

\subsection{Tutorial}

As part of the release of MiniHack, we release an interactive tutorial jupyter notebook for the \desfile{} format, which utilises MiniHack to visualise how the \desfile{} affects the generated level.\footnote{The tutorial can be found in MiniHack's documentation at \url{https://minihack.readthedocs.io}.}
Here we present an overview of the different kinds of \texttt{des-files}, how to add entities to levels, and the main sources of randomness that can be used to create a distribution of levels on which to train RL agents. An in-depth reference can also be found in \cite{des-file}. 

\subsection{Types of \texttt{des-files}}\label{appendix:des-file-types}

There are two types of levels that can be created using \desfile{} format, namely MAZE-type and ROOM-type:
\begin{enumerate}
    \item MAZE-type levels are composed of maps of the level (specified with the \textcolor{blue}{\texttt{MAP}} command) which are drawn using ASCII characters (see \cref{fig:Maze_Des_Example} lines 4-14), followed by descriptions of the contents of the level, described in detail below. In MAZE-type environments, the layout of the map is fixed, but random terrain can be created around (or within) that map using the \textcolor{blue}{\texttt{MAZEWALK}} command, which creates a random maze from a given location and filling all available space of a certain terrain type.
    \item ROOM-type levels are composed of descriptions of rooms (specified by the \textcolor{blue}{\texttt{ROOM}} command), each of which can have its contents specified by the commands described below. Generally, the \textcolor{blue}{\texttt{RANDOM\_CORRIDORS}} command is then used to create random corridors between all the rooms so that they are accessible. On creation, the file specifies (or leaves random) the room's type, lighting and approximate location. It is also possible to create subrooms (using the \textcolor{blue}{\texttt{SUBROOM}} command) which are rooms guaranteed to be within the outer room and are otherwise specified as normal rooms (but with a location relative to the outer room). \cref{fig:oracle} illustrates an instance of the Oracle level specified by the ROOM-type \desfile{} in \cref{fig:Room_Des_Example}.
\end{enumerate}

\subsection{Adding Entities to \texttt{des-files}}\label{appendix:des-file-entities}
As we have seen above, there are multiple ways to define the layout of a level using the \desfile{} format. Once the layout is defined, it is useful to be able to add entities to the level. These could be monsters, objects, traps or other specific terrain features (such as sinks, fountains or altars). In general, the syntax for adding one of these objects is:
\begin{lstlisting}[frame=none]
ENTITY: specification, location, extra-details
\end{lstlisting}
For example:
\begin{lstlisting}[frame=none]
MONSTER: ('F',"lichen"), (1,1)
OBJECT: ('%',"apple"), (10,10)
TRAP: 'fire', (1,1)
# Sinks and Fountains have no specification
SINK: (1,1)
FOUNTAIN: (0,0)
\end{lstlisting}

As can be seen in \cref{fig:Room_Des_Example} lines 4-11, some objects (such as \texttt{statue}s) have extra details that can be specified, such as the monster type (\texttt{montype}) of the statue. Note that many of the details here can instead be set to \texttt{random}, as shown for example in \cref{fig:Room_Des_Example}, lines 29 and 33-36. In this case, the game engine chooses a suitable value for that argument randomly each time the level is generated. For monsters and objects, this randomness can be controlled by just specifying the class of the monster or object and letting the specific object or monster be chosen randomly. For example:
\begin{lstlisting}[frame=none]
MONSTER: 'F', (1,1)
OBJECT: '%', (10,10)
\end{lstlisting}
This code would choose a random monster from the Fungus class (\url{https://nethackwiki.com/wiki/Fungus}), and a random object from the Comestible class (\url{https://nethackwiki.com/wiki/Comestible}).

\subsection{Sources of Randomness in \desfile{}}\label{appendix:des-file-randomness}
We have seen how to create either very fixed (MAZE-type) or very random (ROOM-type) levels, and how to add entities with some degree of randomness.
The \desfile{} format has many other ways of adding randomness, which can be used to control the level generation process, including where to add terrain and in what way. Many of these methods are used in \texttt{IF} statements, which can be in one of two forms:
\begin{lstlisting}[frame=none]
IF[50%] {
	MONSTER: 'F', (1,1)
} ELSE {
	# ELSE is not always necessary
	OBJECT: '%', (1,1)
}

IF[$variable_name < 15] {
	MONSTER: 'F', (1,1)
}
\end{lstlisting}
In the first form, a simple percentage is used for the random choice, whereas in the second, a variable (which could have been randomly determined earlier in the file) is used. A natural way to specify this variable is either in other conditional statements (perhaps you randomly add some number of monsters, and want to count the number of monsters you add such that if there are many monsters, you also add some extra weapons for the agent), or through dice notation. Dice notation is used to specify random expressions which resolve to integers (and hence can be used in any place an integer would be). They are of the form \texttt{NdM}, which means to roll \texttt{N M}-sided dice and sum the result. For example:

\begin{lstlisting}[frame=none]
$roll = 2d6
IF[$roll < 7] {
    MONSTER: random, random
}
\end{lstlisting}

Dice rolls can also be used for accessing arrays, another feature of the \desfile{} format. Arrays are initialised with one or more objects of the same type, and can be indexed with integers (starting at 0), for example:
\begin{lstlisting}[frame=none]
# An array of monster classes
$mon_letters = { 'A', 'L', 'V', 'H' }
# An array of monster names from each monster class respectively
$mon_names = { "Archon", "arch-lich", "vampire lord", "minotaur" }
# The monster to choose
$mon_index = 1d4 - 1
MONSTER:($mon_letters[$mon_index],$mon_names[$mon_index]),(10,18)
\end{lstlisting}

Another way to perform random operations with arrays is using the \textcolor{blue}{\texttt{SHUFFLE}} command. This command takes an array and randomly shuffles it. This would not work with the above example, as the monster name needs to match the monster class (i.e. we could not use \texttt{('A', "minotaur")}). For example:
\begin{lstlisting}[frame=none]
$object = object: { '[',')','*','%' }
SHUFFLE: $object
\end{lstlisting}
Now the \textcolor{purple}{\texttt{\$object}} array will be randomly shuffled. Often, something achievable with shuffling can also be achieved with dice-rolls, but it is simpler to use shuffled arrays rather than dice-rolls (for example, if you wanted to guarantee each of the elements of the array was used exactly once, but randomise the order, it is much easier to just shuffle the array and select them in order rather than try and generate exclusive dice rolls).

\subsection{Random Terrain Placement}\label{appendix:des-file-terrain}
When creating a level, we may want to specify the structure or layout of the level (using a MAZE-type level), but then randomly create the terrain within the level, which will determine accessibility and observability for the agent and monsters in the level. As an example, consider the example \desfile{} in \cref{fig:des_hide_seek}. In this level, we start with an empty 11x9 \textcolor{blue}{\texttt{MAP}} (lines 3-13). We first replace 33\% of the squares with clouds \texttt{C}, and then 25\% with trees \texttt{T} (lines 15,16). To ensure that any corner is accessible from any other, we create two random-walk lines using \texttt{randline} from opposite corners and make all squares on those lines floor (\texttt{.}). To give the agent a helping hand, we choose a random square in the centre of the room with \texttt{rndcoord} (which picks a random coordinate from a selection of coordinates, see lines 19,20) and place an apple there.

Several other methods of randomly creating selections such as \texttt{filter} (randomly remove points from a selection) and \texttt{gradient} (create a selection based on a probability gradient across an area) are described in the NetHack wiki \desfile{} format page \cite{des-file}.

\begin{figure}
\begin{lstlisting}[numbers=left]
MAZE: "mylevel", ' '
GEOMETRY:center,center
MAP
...........
...........
...........
...........
...........
...........
...........
...........
...........
ENDMAP
REGION:(0,0,11,9),lit,"ordinary"
REPLACE_TERRAIN:(0,0,11,9), '.', 'C', 33%
REPLACE_TERRAIN:(0,0,11,9), '.', 'T', 25%
TERRAIN:randline (0,9),(11,0), 5, '.'
TERRAIN:randline (0,0),(11,9), 5, '.'
$center = selection: fillrect (5,5,8,8)
$apple_location = rndcoord $center
OBJECT: ('%', "apple"), $apple_location

$monster = monster: { 'L','N','H','O','D','T' }
SHUFFLE: $monster
$place = { (10,8),(0,8),(10,0) }
SHUFFLE: $place
MONSTER: $monster[0], $place[0], hostile
STAIR:$place[2],down
BRANCH:(0,0,0,0),(1,1,1,1)
\end{lstlisting}
\caption{An example \desfile{} based on the \texttt{HideNSeek} environment.\label{fig:des_hide_seek} \cref{fig:task_hide} presents several instances of environments generated using this \desfile{}.}
\end{figure}

\begin{figure}[H]
  \centering
  \begin{subfigure}[b]{0.49\linewidth}
    \centering\footnotesize
\begin{lstlisting}[frame=none]
LEVEL: "oracle"

ROOM: "ordinary" , lit, (3,3), (center,center), (11,9) {
  OBJECT:('`',"statue"),(0,0),montype:'C',1
  OBJECT:('`',"statue"),(0,8),montype:'C',1
  OBJECT:('`',"statue"),(10,0),montype:'C',1
  OBJECT:('`',"statue"),(10,8),montype:'C',1
  OBJECT:('`',"statue"),(5,1),montype:'C',1
  OBJECT:('`',"statue"),(5,7),montype:'C',1
  OBJECT:('`',"statue"),(2,4),montype:'C',1
  OBJECT:('`',"statue"),(8,4),montype:'C',1

  SUBROOM: "delphi" , lit , (4,3) , (3,3) {
    FOUNTAIN: (0, 1)
    FOUNTAIN: (1, 0)
    FOUNTAIN: (1, 2)
    FOUNTAIN: (2, 1)
    MONSTER: ('@', "Oracle"), (1,1)
    ROOMDOOR: false , nodoor , random, random
  }

  MONSTER: random, random
  MONSTER: random, random

}
\end{lstlisting}
  \end{subfigure}
  \hfill
  \begin{subfigure}[b]{0.46\linewidth}
      \centering\footnotesize
\begin{lstlisting}[frame=none]
ROOM: "ordinary" , random, random, random, random {
  STAIR: random, up
  OBJECT: random,random
}

ROOM: "ordinary" , random, random, random, random {
  STAIR: random, down
  OBJECT: random, random
  TRAP: random, random
  MONSTER: random, random
  MONSTER: random, random
}

ROOM: "ordinary" , random, random, random, random {
  OBJECT: random, random
  OBJECT: random, random
  MONSTER: random, random
}

ROOM: "ordinary" , random, random, random, random {
  OBJECT: random, random
  TRAP: random, random
  MONSTER: random, random
}

ROOM: "ordinary" , random, random, random, random {
  OBJECT: random, random
  TRAP: random, random
  MONSTER: random, random
}

RANDOM_CORRIDORS
  \end{lstlisting}
  \end{subfigure}
  \caption{The ROOM-type des-file for the Oracle level in NetHack.\label{fig:Room_Des_Example}}
\end{figure}

\begin{figure}[H]
\centering
\includegraphics[width=0.92\textwidth]{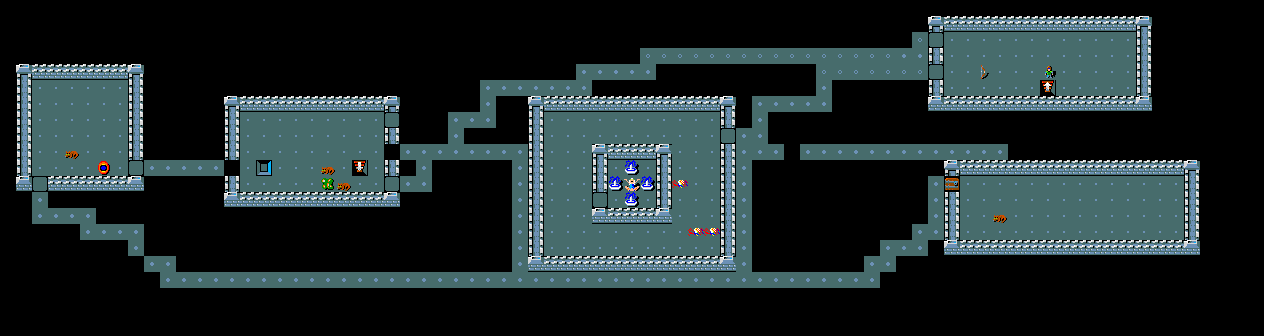}
\caption{A screenshot of the \texttt{Oracle} level specified using the \desfile{} in \cref{fig:Room_Des_Example}.}
\label{fig:oracle}
\end{figure}

\section{MiniHack}\label{appedix:minihack}

\subsection{Observation Space}\label{appendix:obs_space}

MiniHack, like NLE, has a dictionary-structured observation space. Most keys are inherited from NLE, while some are added in MiniHack. Note that using different observation keys can make environments significantly easier or harder (see \cref{appendix:eval_methodology} for discussion of how to ensure comparable experiments are performed).

\begin{itemize}[leftmargin=8pt]
    \item \texttt{glyphs} is a $21\times79$ matrix of glyphs (ids of entities) on the map. Each glyph represents an entirely unique entity, so these are integers between 0 and \texttt{MAX\_GLYPH} (5991). In the standard terminal-based view of NetHack, these glyphs are represented by characters, with colours and other possible visual features.
    \item \texttt{chars} is a $21\times79$ matrix of the characters representing the map.
    \item \texttt{colors} is a $21\times79$ matrix of the colours of each of the characters on the map (some characters represent different objects or monsters depending on their colour).
    \item \texttt{specials} is a $21\times79$ matrix of special extra information about the view of that cell on the map, for example, if the foreground and background colour should be reversed.
    \item \texttt{blstats} ("Bottom Line Stats") is a representation of the status line at the bottom of the screen, containing information about the player character's position, health, attributes and other statuses. It comes in the form of a dimension 25 vector.
    \item \texttt{message} is the utf-8 encoding of the on-screen message displayed at the top of the screen. It's a 256-dimensional vector. 
    \item \texttt{tty\_chars} is the character representation of the entire screen, including the message and map, of size $24\times80$.
    \item \texttt{tty\_colors} is the color representation of the entire screen, including the message and map, of size $24\times80$.
    \item \texttt{tty\_cursor} is the location of the cursor on the screen, a 2-dimensional vector of (x,y) coordinates.
    \item \texttt{inv\_glyphs} is a 55-dimensional vector representing the glyphs present in the current inventory view.
    \item \texttt{inv\_letters} is a 55-dimensional vector representing the letters present in the current inventory view.
    \item \texttt{inv\_oclasses} is a 55-dimensional vector representing the class of objects present in the current inventory view.
    \item \texttt{inv\_strs} is a $55\times80$ matrix containing utf-8 encodings of textual descriptions of objects present in the current inventory view.
\end{itemize}
MiniHack adds the following observation keys:
\begin{itemize}[leftmargin=8pt]
    \item \texttt{screen\_descriptions} is a $21\times79\times80$ tensor of utf-8 encodings of textual descriptions of each cell present in the map. NetHack provides these textual descriptions (which can be accessed by the user by using the describe action on a specific tile).
    \item \texttt{pixel}. We provide pixel-level observations based on the official NetHack tile-set. This is a representation of the current screen in image form, where each cell is represented by a 16x16x3 image, meaning the entire observation is so $336\times1264\times3$ (with 3 channels for RGB). Examples of this pixel observation can be seen in \cref{fig:procgen} and \cref{fig:task_images}.
    \item Cropped observations. For \texttt{glyphs}, \texttt{chars}, \texttt{colors}, \texttt{specials}, \texttt{tty\_chars}, \texttt{tty\_colors}, \texttt{pixel}, and \texttt{screen\_descriptions} a cropped observation centered the agent can be used by passing the observation name suffixed with \texttt{\_crop} (e.g. \texttt{chars\_crop}). This is a NxN matrix centered on the agent's current location containing the information normally present in the full view. The size of the crop can easily be configured using corresponding flags. Cropped observations can facilitate the learning, as the egocentric input makes representation learning easier.
\end{itemize}

\subsection{Interface}\label{appendix:interface}

There are two main MiniHack base classes to chose from, both derived from a common \texttt{MiniHack} base class.

\texttt{MiniHackNavigation} can be used to design mazes and navigation tasks that only require a small action space. All MiniHack navigation tasks we release, as well as MiniGrid and Boxoban examples, make use of the \texttt{MiniHackNavigation} interface. Here the pet is disabled by default. The in-game multiple-choice question prompts as well as menu navigation, are turned off by default.

\texttt{MiniHackSkill} provides a convenient mean for designing diverse skill acquisition tasks that require a large action space and more complex goals. All skill acquisition tasks in MiniHack use this base class. The in-game multiple-choice question prompts is turned on by default, while the menu navigation is turned off. The player's pet and auto-pickup are disabled by default. 

When designing environments, we particularly suggest using partially underspecified levels in order to use the built-in procedural content generator that changes the environment after every episode. For example, several aspects of the level described in \cref{fig:Maze_Des_Example}, such as the types and locations of monsters, objects, and traps, are not fully specified. The NetHack engine will make corresponding selections at random, making that specific feature of the environment procedurally generated. This enables a vast number of environment instances to be created. These could be instances the agent has never seen before, allowing for evaluation of test-time generalisation. 

\subsection{Level Generator}\label{appedix:level_generator}

When creating a new MiniHack environment, a \desfile{} must be provided. One way of providing this \desfile{} is writing it entirely from scratch (for example files see \cref{fig:Room_Des_Example} and  \cref{fig:Maze_Des_Example}, and for example environment creation see \cref{code:des_file}). However, this requires learning the \desfile{} format and is more difficult to do programmatically, so as part of MiniHack we provide the \texttt{LevelGenerator} class which provides a convenient wrapper around writing a \desfile{}. The \texttt{LevelGenerator} class can be used to create MAZE-type levels with specified heights and widths, and can then fill those levels with objects, monsters and terrain, and specify the start point of the level. Combined with the \texttt{RewardManager} which handles rewards (see \cref{appedix:goal_generator}), this enables flexible creation of a wide variety of environments.

The level generator can start with either an empty maze (in which case only height and width are specified, see \cref{code:python} line 2) or with a pre-drawn map (see \cref{code:mixed}). After initialisation, the level generator can be used to add objects, traps, monsters and other terrain features. These can all be either completely or partially specified, in terms of class or location (see \cref{appendix:des-file-entities} and \cref{appendix:des-file-randomness} for more information on how this works, and \cref{code:python} lines 6-10). Terrain can also be added programmatically at a later stage (\cref{code:python} lines 11-12). Once the level is complete, the \texttt{.get\_des()} function returns the \desfile{} which can then be passed to the environment creation function (\cref{code:python} lines 26-29).

\subsection{Reward Manager}\label{appedix:goal_generator}

Along with creating the level layout, structure and content through the \texttt{LevelGenerator}, we also provide an interface to design custom reward functions. The default reward function of MiniHack is a sparse reward of +1 for reaching the staircase down (which terminates the episode), and 0 otherwise, with episodes terminating after a configurable number of time-steps. Using the \texttt{RewardManager}, one can control what events give the agent reward, whether those events can be repeated, and what combinations of events are sufficient or required to terminate the episode.

In \cref{code:python} lines 14-24 present an example of the reward manager usage. We first instantiate the reward manager and add events for eating an apple or wielding a dagger. The default reward is +1, and in general, all events are required to be completed for the episode to terminate. In lines 23-24, we add an event of standing on a sink which has a reward of -1 and is not required for terminating the episode.

While the basic reward manager supports many events by default, users may want to extend this interface to define their own events. This can be done easily by inheriting from the \texttt{Event} class and implementing the \texttt{check} and \texttt{reset} methods. Beyond that, custom reward functions can be added to the reward manager through \texttt{add\_custom\_reward\_fn} method. These functions take the environment instance, the previous observation, action taken and current observation, and should return a float.

We also provide two ways to combine events in a more structured way. The \texttt{SequentialRewardManager} works similarly to the normal reward manager but requires the events to be completed in the sequence they were added, terminating the episode once the last event is complete. The \texttt{GroupedRewardManager} combines other reward managers, with termination conditions defined across the reward managers (rather than individual events). This allows complex conjunctions and disjunctions of groups of events to specify termination. For example, one could specify a reward function that terminates if a sequence of events (a,b,c) was completed, or all events \{d,e,f\} were completed in any order and the sequence (g,h) was completed.

\subsection{Examples}\label{appendix:interface_example}

\begin{figure}
        \centering\footnotesize
        \begin{minipage}{0.48\linewidth}
        \begin{subfigure}{\linewidth}
                \begin{minted}[linenos]{python}
# des_file is the path to the des-file 
# or its content as a string
env = gym.make(
    "MiniHack-Navigation-Custom-v0", 
    def_file=des_file)
            \end{minted}
            \caption{Creating a MiniHack navigation task via \desfile{}.\\\label{code:des_file}}
        \end{subfigure}
        \begin{subfigure}{\textwidth}
                \begin{minted}[linenos]{python}
# Define a 10 by 10 grid area
lvl_gen = LevelGenerator(w=10, h=10)

# Populate it with different objects,
# a monster (goblin) and features (lava)
lvl_gen.add_object("apple", "%")
lvl_gen.add_object("dagger", ")")
lvl_gen.add_trap(name="teleport")
lvl_gen.add_sink()
lvl_gen.add_monster("goblin")
lvl_gen.fill_terrain("rect", "L", 
0, 0, 9, 9)

# Define a reward manager
reward_manager = RewardManager()
# +1 reward and termination for eating 
# an apple or wielding a dagger
reward_manager.add_eat_event("apple")
reward_manager.add_wield_event("dagger")
# -1 reward for standing on a sink
# but isn't required for terminating
# the episode
reward_manager.add_location_event("sink",
reward=-1, terminal_required=False)

env = gym.make(
"MiniHack-Skill-Custom-v0", 
des_file=lvl_gen.get_des(),
reward_manager=reward_manager)
        \end{minted}
            \caption{Creating a skill task using the \texttt{LevelGenerator} and \texttt{RewardManager}.\label{code:python}}
        \end{subfigure}
        \end{minipage}
        \hfil
        \begin{minipage}{0.46\linewidth}
        \begin{subfigure}{\linewidth}
  \begin{minted}{python}
# Define the maze as a string
maze = """
--------------------
|.......|.|........|
|.-----.|.|.-----|.|
|.|...|.|.|......|.|
|.|.|.|.|.|-----.|.|
|.|.|...|....|.|.|.|
|.|.--------.|.|.|.|
|.|..........|...|.|
|.|--------------|.|
|..................|
--------------------
"""
# Set a start and goal positions
lvl_gen = LevelGenerator(map=maze)
lvl_gen.set_start_pos((9, 1))
lvl_gen.add_goal_pos((14, 5))
# Add a Minotaur at fixed position
lvl_gen.add_monster(name="minotaur", 
    place=(19, 9))
# Add wand of death
lvl_gen.add_object("death", "/")

env = gym.make(
    "MiniHack-Skill-Custom-v0", 
    des_file = lvl_gen.get_des())
  \end{minted}
    \caption{Creating a MiniHack skill task using the \texttt{LevelGenerator} with a pre-defined map layout.\label{code:mixed}}
    \end{subfigure}
    \end{minipage}
\caption{Three ways to create MiniHack environments: using only a des-file, using the LevelGenerator and the RewardManager, and LevelGenerator with a pre-defined map layout.}
\label{fig:levelgeneration}
\end{figure}

\cref{code:des_file} presents how to create a MiniHack navigation task using only the \desfile{}, as in \cref{fig:Maze_Des_Example} or \cref{fig:Room_Des_Example}.

\cref{code:python} shows how to create a simple skill acquisition task that challenges the agent to eat an apple and wield a dagger that is randomly placed in a 10x10 room surrounded by lava, alongside a goblin and a teleportation trap. Here, the \texttt{RewardManager} is used to specify the tasks that need to be completed.

\cref{code:mixed} shows how to create a labyrinth task. Here, the agent starts near the entrance of a maze and needs to reach its centre. A Minotaur is placed deep inside the maze, which is a powerful monster capable of instantly killing the agent in melee combat. There is a wand of death placed in a random location in the maze. The agent needs to pick it up, and upon seeing the Minotaur, zap it in the direction of the monster. Once the Minotaur is killed, the agent needs to navigate itself towards the staircase (this is the default goal when \texttt{RewardManager} is not used).
\section{MiniHack tasks}\label{appendix:tasks}

This section presents detailed descriptions of existing MiniHack task and registered configurations. Tasks are grouped into similar tasks, within which several attributes are varied to make more difficult versions of the same task.

\subsection{Navigation Tasks}\label{appendix:nav_tasks}

\paragraph{Room.} 
These tasks are set in a single square room, where the goal is to reach the staircase down (see \cref{fig:task_room}). There are multiple variants of this level. There are two sizes of the room (\texttt{5x5, 15x15}). In the simplest variants, \texttt{Room-5x5} and \texttt{Room-15x15}), the start and goal position are fixed. In the \texttt{Room-Random-5x5} and \texttt{Room-Random-15x15} tasks, the start and goal position are randomised. The rest of the variants add additional complexity to the randomised version of the environment by introducing monsters (\texttt{Room-Monster-5x5} and \texttt{Room-Monster-15x15}), teleportation traps (\texttt{Room-Trap-5x5} and \texttt{Room-Trap-15x15}), darkness (\texttt{Room-Dark-5x5} and \texttt{Room-Dark-15x15}), or all three combined (\texttt{Room-Ultimate-5x5} and \texttt{Room-Ultimate-15x15}).\footnote{The agent can attack monsters by moving towards them when located in an adjacent grid cell. Stepping on a lava tile instantly kills the agent. When the room is dark, the agent can only observe adjacent grid cells.}

\begin{figure}[H]
\centering
\foreach \x in {0,1,2,3,4,5}
{ 
    \includegraphics[width=0.15\textwidth]{figures/screens/room_\x.png}
}
\caption{Various instances of the \texttt{Room-Ultimate-15x15} task.}
\label{fig:task_room}
\end{figure}

\begin{figure}[H]
\centering
\includegraphics[width=0.49\textwidth]{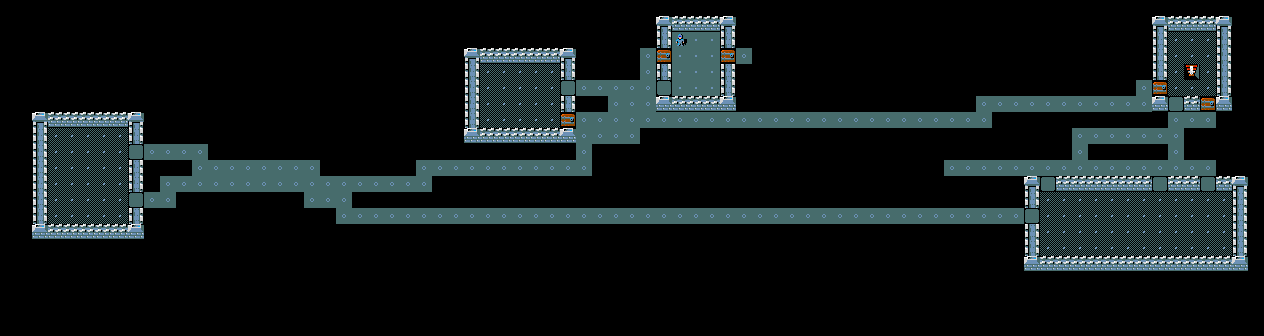}~
\includegraphics[width=0.49\textwidth]{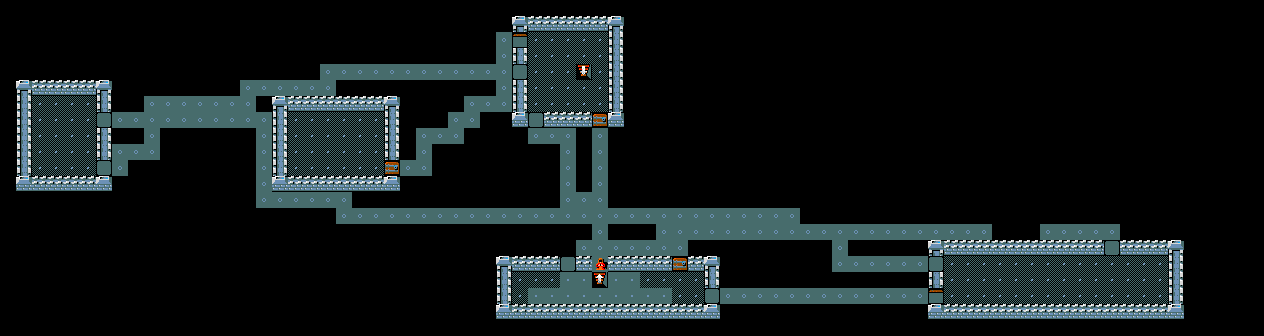}\\
\includegraphics[width=0.49\textwidth]{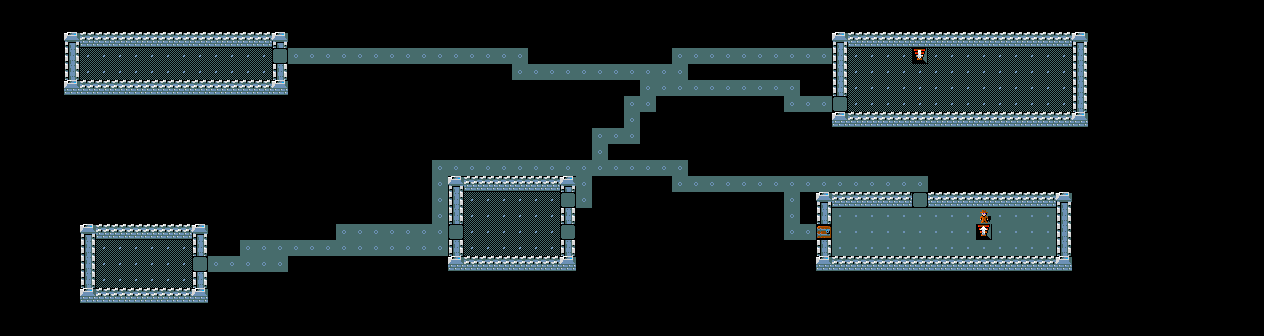}~
\includegraphics[width=0.49\textwidth]{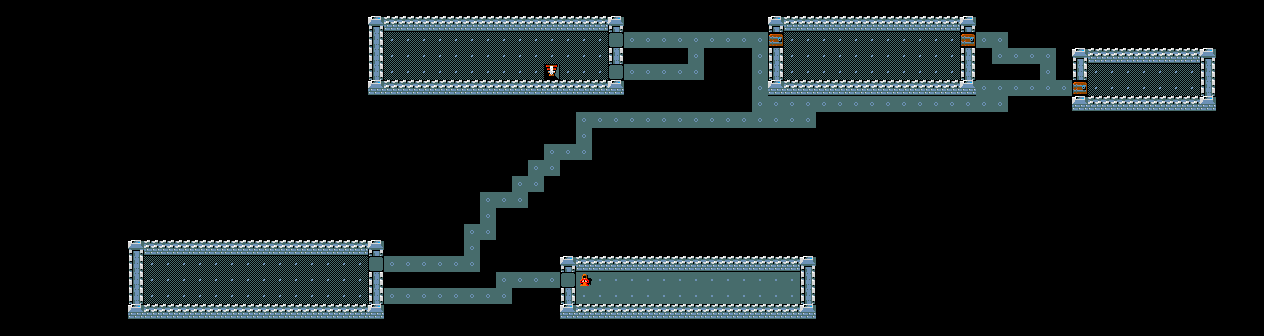}\\
\caption{Four instances of the \texttt{Corridor-R5} task.}
\label{fig:task_corr}
\end{figure}

\paragraph{Corridor.}
These tasks make use of the \texttt{RANDOM\_CORRIDORS} command in the \desfile{}. The objective is to reach the staircase down, which is in a random room (see \cref{fig:task_corr}). The agent is also in a random room. The rooms have randomised positions and sizes. The corridors between the rooms are procedurally generated and are different for every episode. Different variants of this environment have different numbers of rooms, making the exploration challenge more difficult (\texttt{Corridor-R2}, \texttt{Corridor-R3}, and \texttt{Corridor-R5} environments are composed of 2, 3, and 5 rooms, respectively).

\begin{figure}[H]
\centering

\foreach \x in {0,1,2,3,4,5}
{ 
    \includegraphics[width=0.15\textwidth]{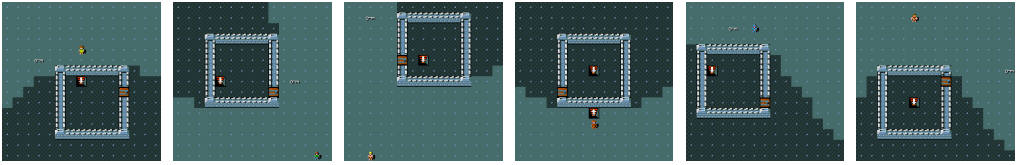}
}
\caption{Various instances of the \texttt{KeyRoom-S15} task.}
\label{fig:task_keyroom}
\end{figure}

\paragraph{KeyRoom.}
These tasks require an agent to pick up a key, navigate to a door, and use the key to unlock the door, reaching the staircase down within the locked room. The action space is the standard movement actions plus the \texttt{PICKUP} and \texttt{APPLY} actions (see \cref{fig:task_keyroom}). In the simplest variant of this task, (\texttt{KeyRoom-Fixed-S5}), the location of the key, door and staircase are fixed. In the rest of the variants, these locations randomised. The size the outer room is 5x5 for \texttt{KeyRoom-S5} and 15x15 for \texttt{KeyRoom-S15}. To increase the difficulty of the tasks, dark versions of the tasks are introduced (\texttt{KeyRoom-Dark-S5} and \texttt{KeyRoom-Dark-S15}), where the key cannot be seen if it is not in any of the agent's adjacent grid cells.

\paragraph{MazeWalk.} 
These navigation tasks make use of the \texttt{MAZEWALK} command in the \desfile{}, which procedurally generates diverse mazes on the 9x9, 15x15 and 45x19 grids for \texttt{MazeWalk-9x9}, \texttt{MazeWalk-15x15}, and \texttt{MazeWalk-45x19} environments, respectively (see \cref{fig:task_mzwk}). In the mapped versions of these tasks (\texttt{MazeWalk-Mapped-9x9}, \texttt{MazeWalk-Mapped-15x15}, and \texttt{MazeWalk-Mapped-45x19}), the map of the maze and the goal's locations are visible to the agent.

\begin{figure}[H]
\centering
\foreach \x in {0,1,2,3,4,5,6,7}
{ 
     \includegraphics[width=0.1132\textwidth]{figures/screens/mzwk_\x.png}
}
\caption{Various instances of the \texttt{MazeWalk-15x15} task.}
\label{fig:task_mzwk}
\end{figure}

\begin{figure}[H]
\centering
\foreach \x in {0,1,2}
{ 
     \includegraphics[width=0.32\textwidth]{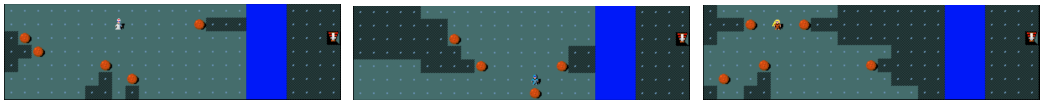}
}
\caption{Three instances of the \texttt{River} task.}
\label{fig:task_river}
\end{figure}

\paragraph{River.} This group of tasks requires the agent to cross a river using boulders (see \cref{fig:task_river}). Boulders, when pushed into the water, create a dry land to walk on, allowing the agent to cross it. While the \texttt{River-Narrow} task can be solved by pushing one boulder into the water, other \texttt{River} require the agent to plan a sequence of at least two boulder pushes into the river next to each other. In the more challenging tasks of the group, the agent needs to additionally fight monsters (\texttt{River-Monster}), avoid pushing boulders into lava rather than water (\texttt{River-Lava}), or both (\texttt{River-MonsterLava}).

\begin{figure}[H]
\centering
\foreach \x in {0,1,2,3}
{ 
     \includegraphics[height=2.2cm]{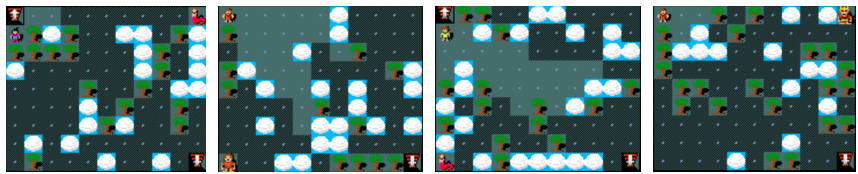}
}
\caption{Four instances of the \texttt{HideNSeek} task.}
\label{fig:task_hide}
\end{figure}

\paragraph{HideNSeek.} In the \texttt{HideNSeek} task, the agent is spawned in a big room full of trees and clouds (see  \cref{fig:task_hide}). The trees and clouds block the line of sight of the player and a random monster (chosen to be more powerful than the agent). The agent, monsters and spells can pass through clouds unobstructed. The agent and monster cannot pass through trees. The goal is to make use of the environment features, avoid being seen by the monster and quickly run towards the goal. The layout of the map is procedurally generated, hence requires systematic generalisation. Alternative versions of this task additionally include lava tiles that need to be avoided (\texttt{HideNSeek-Lava}), have bigger size (\texttt{HideNSeek-Big}) or provide the locations of all environment features but not the powerful monster (\texttt{HideNSeek-Mapped}).

\begin{figure}[H]
\centering
     \includegraphics[width=0.63\textwidth]{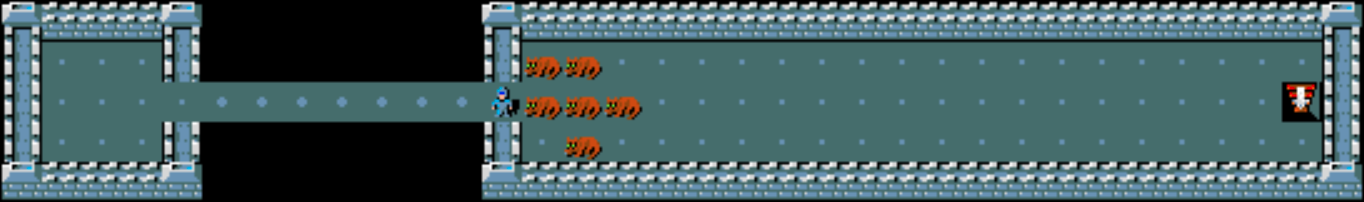}
\caption{A screenshot of the \texttt{CorridorBattle} task.}
\label{fig:task_battle}
\end{figure}

\paragraph{CorridorBattle.} The \texttt{CorridorBattle} task challenges the agent to make best use of the dungeon features to effectively defeat a horde of hostile monsters (see \cref{fig:task_battle}). Here, if the agent lures the rats into the narrow corridor, it can defeat them one at a time. Fighting in rooms, on the other hand, would result in the agent simultaneously incurring damage from several directions and quick death. The task also is offered in dark mode (\texttt{CorridorBattle-Dark}), challenging the agent to remember the number of rats killed in order to plan subsequent actions.

\begin{figure}[H]
\centering
     \includegraphics[width=0.92\textwidth]{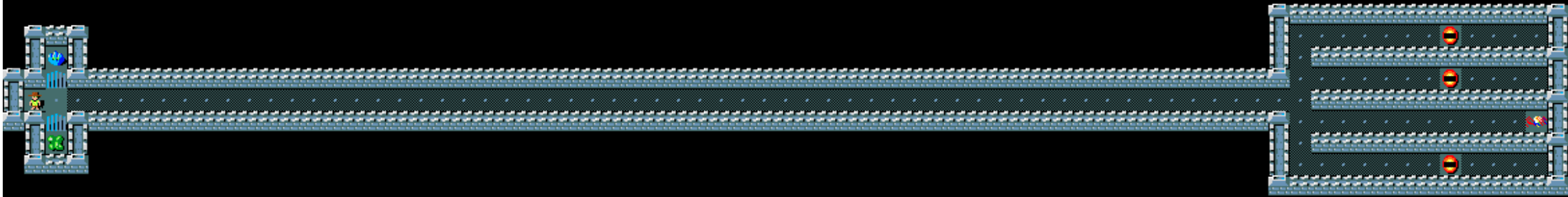}
\caption{A screenshot of the \texttt{Memento-F4} task.}
\label{fig:task_memento}
\end{figure}

\paragraph{Memento.} This group of tasks test the agent's ability to use memory (within an episode) to pick the correct path. The agent is presented with a prompt (in the form of a sleeping monster of a specific type) and then navigates along a corridor (see \cref{fig:task_memento}). At the end of the corridor, the agent reaches a fork and must choose a direction. One direction leads to a grid bug, which if killed terminates the episode with a +1 reward. All other directions lead to failure through an invisible trap that terminates the episode when activated. The correct path is determined by the cue seen at the beginning of the episode. We provide three versions of this environment: one with a short corridor before a fork with two paths to choose from (\texttt{Memento-Short-F2}), one with a long corridor with a two-path fork (\texttt{Memento-F2}), and one with a long corridor and a four-fork path (\texttt{Memento-F4}).

\begin{figure}[H]
\centering
\foreach \x in {1,2,3,4}
{ 
     \includegraphics[height=1.59cm]{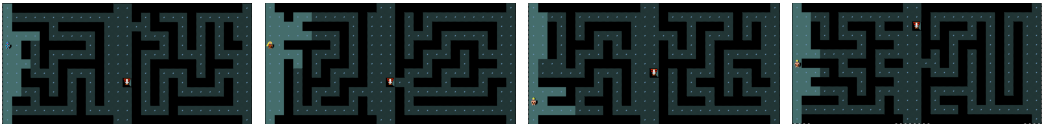}
}
\caption{Four instances of the \texttt{MazeExplore-Hard} task. The apples are located near the right vertical wall (unobservable in the figure). The goal is located in the middle area of the grid.}
\label{fig:task_explr}
\end{figure}

\paragraph{MazeExplore.} These tasks test the agent's ability to perform deep exploration \cite{osband2016deep}. It's inspired by the Apple-Gold domain from \cite{guo2020memory}, where a small reward can be achieved easily, but to learn the optimal policy deeper exploration is required. The agent must first explore a simple randomised maze to reach the staircase down, which they can take for +1 reward (see \cref{fig:task_explr}). However, if they navigate through a further randomised maze, they reach a room with apples. Eating the apples gives a +0.5 reward, and once the apples are eaten the agent should then return to the staircase down. We provide an easy and a hard version of this task (\texttt{MazeExplore-Easy} and \texttt{MazeExplore-Hard}), with the harder version having a larger maze both before and after the staircase down. Variants can also be mapped (\texttt{MazeExplore-Easy-Mapped} and \texttt{MazeExplore-Hard-Mapped}), where the agent can observe the layout of the entire grid, making it easier to navigate the maze. Even in the mapped setting, the apples are not visible until the agent reaches the final room.

The full list of navigation tasks in MiniHack is provided in 
\cref{tab:nav_tasks}.

\begin{table*}
    \centering
	\caption{Full list of MiniHack navigation tasks and corresponding capabilities for assessment.}
	\begin{tabular}{lc}
		\toprule
		Name & Capability\\
		\midrule
		\texttt{Room-5x5-v0}& Basic Learning\\
		\texttt{Room-15x15-v0}& Basic Learning\\
		\texttt{Room-Random-5x5-v0} &  Basic Learning\\
		\texttt{Room-Random-15x15-v0} &  Basic Learning\\
		\texttt{Room-Dark-5x5-v0} & Basic Learning\\
		\texttt{Room-Dark-15x15-v0} & Basic Learning\\
		\texttt{Room-Monster-5x5-v0} & Basic Learning\\
		\texttt{Room-Monster-15x15-v0} & Basic Learning\\
		\texttt{Room-Trap-5x5-v0} & Basic Learning\\
		\texttt{Room-Trap-15x15-v0} & Basic Learning\\
		\texttt{Room-Ultimate-5x5-v0} & Basic Learning\\
		\texttt{Room-Ultimate-15x15-v0} & Basic Learning\\
		\midrule
		\texttt{Corridor-R2-v0} & Exploration \\
		\texttt{Corridor-R3-v0} & Exploration \\
		\texttt{Corridor-R5-v0} & Exploration \\
		\midrule
		\texttt{KeyRoom-Fixed-S5-v0}  & Exploration \\
		\texttt{KeyRoom-S5-v0}  & Exploration \\
		\texttt{KeyRoom-Dark-S5-v0}  & Exploration \\
		\texttt{KeyRoom-S15-v0}  & Exploration \\
		\texttt{KeyRoom-Dark-S15-v0}  & Exploration \\
		\midrule
		\texttt{MazeWalk-9x9-v0} & Exploration \& Memory \\
		\texttt{MazeWalk-Mapped-9x9-v0} & Exploration \& Memory \\
		\texttt{MazeWalk-15x15-v0} & Exploration \& Memory \\
		\texttt{MazeWalk-Mapped-15x15-v0} & Exploration \& Memory \\
		\texttt{MazeWalk-45x19-v0} & Exploration \& Memory \\
		\texttt{MazeWalk-Mapped-45x19-v0} & Exploration \& Memory \\
		\midrule
		\texttt{River-Narrow-v0}  & Planning \\
		\texttt{River-v0}  & Planning \\
		\texttt{River-Monster-v0}  & Planning \\
		\texttt{River-Lava-v0}  & Planning \\
		\texttt{River-MonsterLava-v0}  & Planning \\
		\midrule
		\texttt{HideNSeek-v0} & Planning \\
		\texttt{HideNSeek-Mapped-v0} & Planning \\
		\texttt{HideNSeek-Lava-v0} & Planning \\
		\texttt{HideNSeek-Big-v0} & Planning \\
		\midrule
		\texttt{CorridorBattle-v0} & Planning \& Memory\\
		\texttt{CorridorBattle-Dark-v0} & Planning \& Memory\\
		\midrule
		\texttt{Memento-Short-F2-v0} & Memory\\
		\texttt{Memento-F2-v0} & Memory\\
		\texttt{Memento-F4-v0} & Memory\\
		\midrule
		\texttt{MazeExplore-Easy-v0} & Deep Exploration\\
		\texttt{MazeExplore-Hard-v0} & Deep Exploration\\
		\texttt{MazeExplore-Easy-Mapped-v0} & Deep Exploration\\
		\texttt{MazeExplore-Hard-Mapped-v0} & Deep Exploration\\
		\bottomrule
	\end{tabular}
	\label{tab:nav_tasks}
\end{table*}

\subsection{Skill Acquisition Tasks}\label{appendix:skill_tasks}

\subsubsection{Skills}

The nature of commands in NetHack requires the agent to perform a sequence of actions so that the initial action, which is meant for interaction with an object, has an effect. 
The exact sequence of subsequent can be inferred by the in-game message bar prompts.\footnote{Hence the messages are also used as part of observations in the skill acquisition tasks.} For example, when located in the same grid with an apple lying on the floor, choosing the \texttt{Eat} action will not be enough for the agent to eat it. In this case, the message bar will ask the following question: \textit{"There is an apple here; eat it? [ynq] (n)}". Choosing the \texttt{Y} action at the next timestep will cause the initial \texttt{EAT} action to take effect, and the agent will eat the apple. Choosing the \texttt{N} action (or \texttt{MORE} action since \texttt{N} is the default choice) will decline the previous \texttt{EAT} action prompt. The rest of the actions will not progress the in-game timer and the agent will stay in the same state. We refer to this skill as \texttt{Confirmation}.

The \texttt{PickUp} skill requires picking up objects from the floor first and put in the inventory. The tasks with \texttt{InventorySelect} skill necessities selecting an object from the inventory using the corresponding key, for example, \textit{"What do you want to wear? [fg or ?*]"} or \textit{"What do you want to zap? [f or ?*]"}. The \texttt{Direction} skill requires choosing one of the moving directions for applying the selected action, e.g., kicking or zapping certain types of wands. In this case, \textit{"In what direction?"} message will appear on the screen. The \texttt{Navigation} skill tests the agent's ability to solve various mazes and labyrinths using the moving commands.

\subsubsection{Tasks}

The full list of skill acquisition tasks, alongside the skills they require mastering, is provided in 
\cref{tab:skill_tasks}. The skill acquisition tasks are suitable testbeds for fields such as curriculum learning and transfer learning, either between different tasks within MiniHack or to the full game of NetHack. 

\begin{figure}[H]
\centering
\includegraphics[width=0.2\textwidth]{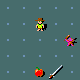}~~~~
\includegraphics[width=0.2\textwidth]{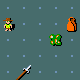}~~~~
\includegraphics[width=0.2\textwidth]{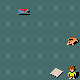}
\caption{Random instances of the \texttt{Eat-Distract}, \texttt{Wear-Distract} and \texttt{Pray-Distract} tasks.}
\label{fig:task_skill_simple}
\end{figure}

\paragraph{Simple Tasks.} The simplest skill acquisition tasks require discovering interaction between one object and the actions of the agent. These include: eating comestibles (\texttt{Eat}), praying on an altar (\texttt{Pray}), wearing armour (\texttt{Wear}), and kicking locked doors (\texttt{LockedDoors}). In the regular versions of these tasks, the starting location of the objects and the agent is randomised, whereas in the fixed versions of these tasks (\texttt{Eat-Fixed}, \texttt{Pray-Fixed}, \texttt{Wear-Fixed} and \texttt{LockedDoors-Fixed}) both are fixed.
To add a slight complexity to the randomised version of these tasks, distractions in the form of a random object and a random monster are added to the third version of these tasks (\texttt{Eat-Distract}, \texttt{Pray-Distract} and \texttt{Wear-Distract}, see \cref{fig:task_skill_simple}).
These tasks can be used as building blocks for more advanced skill acquisition tasks. 

\begin{figure}[H]
\centering
\includegraphics[width=0.185\textwidth]{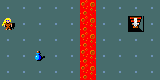}~
\includegraphics[width=0.185\textwidth]{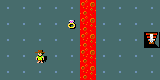}~
\includegraphics[width=0.185\textwidth]{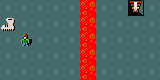}~
\includegraphics[width=0.185\textwidth]{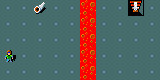}~
\includegraphics[width=0.185\textwidth]{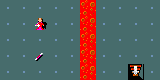}
\caption{Five random instances of the \texttt{LavaCross} task, where the agent needs to cross the lava using (i) potion of levitation, (ii) ring of levitation, (iii) levitation boots, (iv) frost horn, or (v) wand of cold.}
\label{fig:task_skill_lava}
\end{figure}

\paragraph{Lava Crossing.}
An example of a more advanced task involves crossing a river of lava. The agent can accomplish this by either levitating over it (via a potion of levitation or levitation boots) or freezing it (by zapping the wand of cold or playing the frost horn). In the simplest version of the task  (\texttt{LavaCross-Levitate-Potion-Inv} and \texttt{LavaCross-Levitate-Ring-Inv}), the agent starts with one of the necessary objects in the inventory. Requiring the agent to pick up the corresponding object first makes the tasks more challenging (\texttt{LavaCross-Levitate-Potion-PickUp} and \texttt{LavaCross-Levitate-Ring-PickUp}). The most difficult variants of this task group require the agent to cross the lava river using one of the appropriate objects randomly sampled and placed at a random location. In \texttt{LavaCross-Levitate}, one of the objects of levitation is placed on the map, while in the \texttt{LavaCross} task these include all of the objects for levitation as well as freezing (see \cref{fig:task_skill_lava}).

\begin{figure}[H]
\centering
\includegraphics[width=0.49\textwidth]{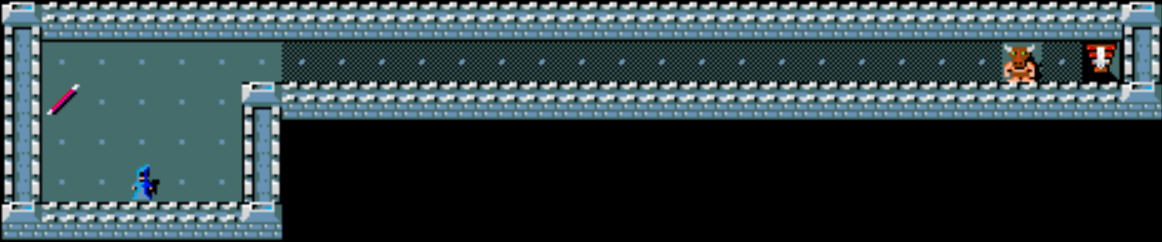}
\caption{A screenshot of the \texttt{WoD-Hard} task.}
\label{fig:task_wod}
\end{figure}

\paragraph{Wand of Death.} MiniHack is very convenient for making incremental changes to the difficulty of a task. To illustrate this, we provide a sequence of tasks that require mastering the usage of the wand of death \citep[WoD, ][]{nhwiki}. Zapping a WoD in any direction fires a death ray which instantly kills almost any monster it hits. In \texttt{WoD-Easy} environment, the agent starts with a WoD in its inventory and needs to zap it towards a sleeping monster. \texttt{WoD-Medium} requires the agent to pick it up, approach the sleeping monster, kill it, and go to the staircase.
In \texttt{WoD-Hard} the WoD needs to be found first, only then the agent should enter the corridor with a monster (who is awake and hostile this time), kill it, and go to the staircase (see \cref{fig:task_wod}). 
In the most difficult task of the sequence, the \texttt{WoD-Pro}, the agent starts inside a big labyrinth. It needs to find the WoD inside the maze and reach its centre, which is guarded by a deadly Minotaur.

\begin{figure}[H]
\centering
\includegraphics[width=0.49\textwidth]{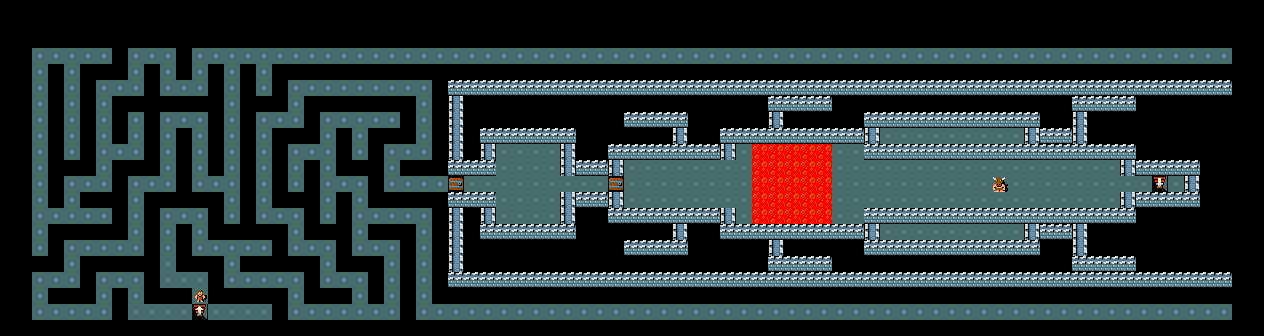}~
\includegraphics[width=0.49\textwidth]{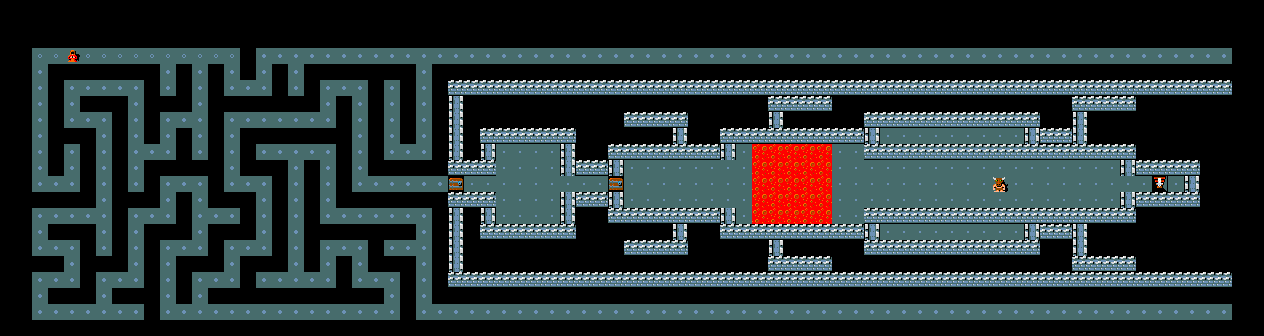}\\
\caption{Two instances of the \texttt{Quest-Hard} task.}
\label{fig:task_quest}
\end{figure}

\paragraph{Quest.} \texttt{Quest} tasks require the agent to navigate mazes, cross rivers and fight powerful monsters. \texttt{Quest\_Easy}, the simplest environment in the group, challenges the agent to use objects in the inventory to cross the river of lava and fight a few relatively weak monsters. In \texttt{Quest\_Medium} the agent engages with a horde of monsters instead and must lure them into the narrow corridors to survive. The \texttt{Quest\_Hard} task, the most difficult environment of the group, requires the agent to solve complex, procedurally generated mazes, find objects for crossing the lava river and make use of the wand of death to kill a powerful monster (see \cref{fig:task_quest}).

\begin{table*}
    \centering
	\caption{Full list of MiniHack skill acquisition tasks.}
	\begin{tabular}{lc}
		\toprule
		Name & Skill \\
		\midrule
		\texttt{Eat-v0} & Confirmation or PickUp+Inventory \\
		\texttt{Eat-Fixed-v0} & Confirmation or PickUp+Inventory \\
		\texttt{Eat-Distract-v0} & Confirmation or PickUp+Inventory \\
		\texttt{Pray-v0} & Confirmation \\
		\texttt{Pray-Fixed-v0} & Confirmation \\
		\texttt{Pray-Distract-v0} & Confirmation \\
		\texttt{Wear-v0} & PickUp+Inventory \\
		\texttt{Wear-Fixed-v0} & PickUp+Inventory \\
		\texttt{Wear-Distract-v0} & PickUp+Inventory \\
		\texttt{LockedDoor-v0} & Direction \\
		\texttt{LockedDoor-Random-v0} & Direction \\
		\midrule
		\texttt{LavaCross-Levitate-Ring-Inv-v0} & Inventory \\
		\texttt{LavaCross-Levitate-Potion-Inv-v0} & Inventory \\
		\texttt{LavaCross-Levitate-Ring-Pickup-v0} & PickUp+Inventory \\
		\texttt{LavaCross-Levitate-Potion-PickUp-v0} & PickUp+Inventory \\
		\texttt{LavaCross-Levitate-v0} & PickUp+Inventory \\
		\texttt{LavaCross-v0} & PickUp+Inventory \\
		\midrule
		\texttt{WoD-Easy} & Inventory+Direction \\
		\texttt{WoD-Medium} & PickUp+Inventory+Direction \\
		\texttt{WoD-Hard} & PickUp+Inventory+Direction \\
		\texttt{WoD-Pro} & Navigation+PickUp+Inventory+Direction \\
		\midrule
		\texttt{Quest-Easy-v0} & Inventory \\
		\texttt{Quest-Medium-v0} & Navigation+Inventory \\
		\texttt{Quest-Hard-v0} & Navigation+PickUp+Inventory+Direction \\
		\bottomrule
	\end{tabular}
	\label{tab:skill_tasks}
\end{table*}

\subsection{Ported Tasks}\label{appendix:ported}

The full list of tasks ported to MiniHack from MiniGrid~\cite{gym_minigrid} and Boxoban~\cite{boxobanlevels} which we used in our experiments is provided in \cref{tab:ported_tasks}. Note that more tasks could have similarly been ported from MiniGrid. However, our goal is to showcase MiniHack's ability to port existing gridworld environments and easily enrich them, rather than porting all possible tasks.

\begin{table*}
    \centering
	\caption{Tasks ported to MiniHack from other benchmarks.}
	\begin{tabular}{lc}
		\toprule
		Name & Capability \\
		\midrule
		\texttt{MultiRoom-N2-v0} \cite{gym_minigrid} & Exploration \\
		\texttt{MultiRoom-N4-v0} \cite{gym_minigrid} & Exploration \\
		\texttt{MultiRoom-N2-Monster-v0} & Exploration \\
		\texttt{MultiRoom-N4-Monster-v0} & Exploration \\
		\texttt{MultiRoom-N2-Locked-v0} & Exploration \\
		\texttt{MultiRoom-N4-Locked-v0} & Exploration \\
		\texttt{MultiRoom-N2-Lava-v0} & Exploration \\
		\texttt{MultiRoom-N4-Lava-v0} & Exploration \\
		\texttt{MultiRoom-N2-Extreme-v0} & Exploration \\
		\texttt{MultiRoom-N4-Extreme-v0} & Exploration \\
		\midrule
		\texttt{Boxoban-Unfiltered-v0} \cite{boxobanlevels} & Planning \\
		\texttt{Boxoban-Medium-v0} \cite{boxobanlevels} & Planning \\
		\texttt{Boxoban-Hard-v0} \cite{boxobanlevels} & Planning \\
		\bottomrule
	\end{tabular}
	\label{tab:ported_tasks}
\end{table*}

\section{Experiment Details}\label{appendix:experiments}

Instructions on how to replicate experiments we present can be found in MiniHack's repository: \url{https://github.com/facebookresearch/minihack}.
Below we provide details on individual components.

\subsection{Agent and Environment Details}\label{appendix:agent}

The agent architecture used throughout all experiments is identical to it in \cite{kuttler2020nethack}.
The observations used by the model include the $21\times 79$ matrix of grid entity representations and a
$21$-dimensional vector containing agent statistics, such as its coordinates, health points, etc.
Every of the $5991$ possible entities in NetHack (monsters, items, dungeon features, etc.) is mapped onto a $k$-dimensional vector representation as follows. First, we split the entity ids (glyphs) into one of twelve groups (categories of entities) and an id within each group. We construct a partition of the final vector which includes the following components as sub-vectors: group, subgroup\_id, color, character, and special, each of which uses a separate embedding that is learned throughout the training. The relative length of each sub-vector is defined as follows: groups=1, subgroup\_ids=3, colors=1, chars=2, and specials=1. That is, for a $k=64$ dimensional embeddings, we make use of an embedding dimension of 24 for the id, 8 for group, 8 for color, 16 for character, and 8 for special. These settings were determined during a set of small-scale experiments.

Three dense representations are produced.
First, all visible entity embeddings are passed to a CNN. 
Second, another CNN is applied to the $9\times 9$ crop of entities surrounding the agent.
Third, an MLP is used to encode the agent's statistic.  
These three vectors are concatenated and passed to another MLP which produces the final representation $\mathbf{o}_t$ of the observation. 
To obtain the action distribution, we feed the observations $\mathbf{o}_t$ to a recurrent layer comprised with an LSTM~\cite{DBLP:journals/neco/HochreiterS97} cells, followed by an additional MLP.

For results on skill acquisition tasks, the in-game message, encoded using a character-level CNN~\cite{zhangZL15}, is also included as part of observation.

For all conducted experiments, a penalty of $-0.001$ is added to the reward function if the selected action of the agent
does not increment the in-game timer of NetHack. For instance, when the agent attempts to move against a wall or navigates in-game menus, it will receive the $-0.001$ penalty.

\subsection{TorchBeast Details}\label{appendix:torchbeast}

We employ an embedding dimension of $64$ for entity representations. The size of the hidden dimension for the observation $\mathbf{o}_t$ and the output of the LSTM $\mathbf{h}_t$ is $256$.
We use a $5$-layer CNN architecture (filter size $3\times3$, padding $1$, stride $1$) for encoding both the full screen of entities and the $9\times 9$ agent-centred crop.
The input channel of the first layer of the CNN is the embedding size of entities ($64$). 
The dimensions of all subsequent layers are equal to $16$ for both input and output channels. 

The characters within the in-game messages are encoded using an embedding of size $32$ and passed to a $6$-layer CNN architecture, each comprised of a 1D convolution of size $64$ and ReLU nonlinearity. Maxpooling is applied after the first, second, and sixth layers. The convolutional layers are followed by an MLP.

We apply a gradient norm clipping of $40$, but do not clip the rewards. 
We employ an RMSProp optimiser with a learning rate of $2*10^{-4}$ without momentum and
with $\epsilon = 10^{-6}$.\footnote{For results on skill acquisition tasks, we use a learning rate of $5*10^{-5}$.} The entropy cost is set to $10^{-4}$. The $\gamma$ discounting factor is set to $0.999$.

The hyperparameters for Random Network Distillation \citep[RND,][]{ebner_towards_2013} are the same as in \citep{kuttler2020nethack} and mostly follow the author recommendations.
The weights are initialised using an orthogonal distribution with gains of $\sqrt{2}$. A two-headed value function is used for intrinsic and extrinsic rewards. The discount factor for the intrinsic reward is set to $0.99$.
We treat both extrinsic and intrinsic rewards as episodic in nature.
The intrinsic reward is normalised by dividing it by a running estimate of its standard deviation.
Unlike the original implementation of RND, we do not use observation normalisation due to the symbolic nature of observations used in our experiments.
The intrinsic reward coefficient is $0.1$. Intrinsic rewards are not clipped.

For our RIDE \cite{raileanu2020ride} baselines, we normalise the intrinsic reward by the number of visits to a state. The intrinsic reward coefficient is $0.1$. The forward and inverse dynamics models have hidden dimension of $128$. The loss cost is $1$ for the forward model and $0.1$ for inverse model.

These settings were determined during a set of small-scale experiments.

The training on MiniHack's \texttt{Room-5x5} task for two million timesteps using our IMPALA baseline takes approximately 4:30 minutes (roughly $7667$ steps per second). This estimate is measured using 2 NVIDIA Quadro GP100 GPUs (one for the learner and one for actors) and 20 Intel(R) Xeon(R) E5-2698 v4 @ 2.20GHz CPUs (used by 256 simultaneous actors) on an internal cluster. Our RND baseline completes the same number of timesteps in approximately 7:30 minutes (roughly $4092$ steps per second) using the same computational resources. 

\subsection{Agent Architecture Comparison Details}\label{appendix:arch_comparison}

Here we provide details on the architectures of models used in \cref{fig:arch_comparison}. The \textit{medium} model uses the exact architectures described in \cref{appendix:torchbeast}, namely a $5$-layer CNN, hidden dimension size $256$, and entity embedding size $64$. The \textit{small} model uses a $3$-layer CNN, hidden dimension size $64$, and entity embedding size $16$. The \textit{large} model uses a $9$-layer CNN, hidden dimension size $512$, and entity embedding size $128$. The rest of the hyperparameters are identical for all models and are described in \cref{fig:arch_comparison}.

\subsection{RLlib Details}\label{appendix:rllib}

We release examples of training several RL algorithms on MiniHack using RLlib \cite{pmlr-v80-liang18b}. RLlib is an open-source scalable reinforcement learning library built on top of Ray \cite{moritz2018ray}, that provides a unified framework for running experiments with many different algorithms. We use the same model as in the TorchBeast implementation described above, adjusted to not manage the time dimension for the models that use an LSTM (as RLlib handles the recurrent network logic separately). To enable future research on a variety of methods, we provide examples of training DQN~\cite{mnih2015human}, PPO~\cite{Schulman2017ProximalPO} and A2C~\cite{mnih2016asynchronous} on several simple MiniHack environments. Our DQN agent makes use of Double Q-Learning \cite{van2016deep}, duelling architecture \cite{wang2016dueling} and prioritized experience replay (PER) \cite{schaul2015prioritized}. We perform a limited sweep over hyperparameters, as we are not trying to achieve state-of-the-art results, just provide a starting point for future research.

In all experiments, we use 1 GPU for learning, and 10 CPUs for acting (with multiple instances of each environment per CPU), to speed up the wall-clock run-time of the experiments. These options can be configured easily in the code we release.

Results for these experiments can be seen in \cref{fig:rllib}. Hyperparameters for DQN, PPO, and A2C are detailed in \cref{tab:rllib_hp_dqn},  \cref{tab:rllib_hp_ppo} and \cref{tab:rllib_hp_a2c}, respectively.

\begin{table}[H]
    \centering
    \caption{RLlib DQN Hyperparameters}\label{tab:rllib_hp_dqn}
    \begin{tabular}{lc}
        \toprule
        Name & value \\
        \midrule
         learning rate & 1e-6 \\
         replay buffer size & 100000 \\
         PER $\beta$ & 0.4 \\
         n-step length & 5 \\
         target network update frequency & 50000 \\
         learning start steps & 50000 \\
         PER annealing timesteps & 100000 \\
         \bottomrule
    \end{tabular}
\end{table}

\begin{table}[H]
\centering
    \caption{RLlib PPO Hyperparameters}\label{tab:rllib_hp_ppo}
    \begin{tabular}{lc}
        \toprule
        Name & value \\
        \midrule
         learning rate & 1e-5 \\
         batch size & 128 \\
         SGD minibatch size & 32 \\
         SGD iterations per epoch & 2 \\
         rollout fragment length & 128 \\
         entropy penalty coefficient & 0.0001 \\
         value function loss coefficient & 0.5 \\
         shared policy and value representation & True \\
         \bottomrule
    \end{tabular}
\end{table}

\begin{table}[H]
\centering 
    \caption{RLlib A2C Hyperparameters}\label{tab:rllib_hp_a2c}
    \begin{tabular}{lc}
        \toprule
        Name & value \\
        \midrule
         learning rate & 1e-5 \\
         batch size & 128 \\
         rollout fragment length & 128 \\
         entropy penalty coefficient & 0.001 \\
         value function loss coefficient & 0.1 \\
         \bottomrule
    \end{tabular}
\end{table}

\subsection{Unsupervised Environment Design}\label{appendix:ued}

We base our UED experiments using PAIRED on MiniHack largely on those outlined in \citep{dennis2020emergent}. The hyperparameters used for training are provided in \cref{tab:hp_ued}. Note that except where explicitly noted, all agents share the same training hyperparameters.

The adversary constructs new levels starting from an empty $5 \times 5$ grid. At each of the first 10 timesteps, the adversary chooses a position in which to place one of the following objects: \{\texttt{walls}, \texttt{lava}, \texttt{monster}, \texttt{locked door}\}. If a selected cell is already occupied, no additional object cell is placed. After placing the objects, the adversary then chooses the goal position followed by the agent's starting position. If a selected cell is already occupied, the position is randomly resampled from among the free cells.

At each time step, the adversary policy encodes the glyph observation using two convolution layers, each with kernel size $3\times3$, stride lengths of 1 and 2, and output channels, 16 and 32 respectively, followed by a ReLU activation over the flattened outputs. We embed the time step into a 10-dimensional space. The image embedding, time-step embedding, and the random noise vector are concatenated, and the combined representation is passed through an LSTM with a hidden dimension of 256, followed by two fully connected layers with a hidden dimension of 32 and ReLU activations to yield the action logits over the 169 possible cell choices. 

We make use of the same architecture for the protagonist and antagonist policies, with the exceptions of using the agent-centred crop, rather than the full glyph observation, and producing policy logits over the MiniHack action space rather than over the set of possible cell positions. 

\begin{table}[H]
\centering 
    \caption{PAIRED hyperparameters\label{tab:hp_ued}}
    \begin{tabular}{lc}
        \toprule
        Name & value \\
        \midrule
		$\gamma$ & 0.995 \\
		$\lambda_{GAE}$ & 0.95 \\
		PPO rollout length & 256  \\
		PPO epochs & 5 \\
		PPO minibatches per epoch & 1 \\
		PPO clip range & 0.2 \\
		PPO number of workers & 32 \\
		Adam learning rate & 1e-4 \\
		Adam $\epsilon$ & 1e-5 \\
		PPO max gradient norm & 0.5 \\
		PPO value clipping & yes \\
		value loss coefficient & 0.5 \\
		protagonist/antagonist entropy coefficient & 0.0 \\
		adversary entropy coefficient & 0.005 \\
         \bottomrule
    \end{tabular}
\end{table}

\section{Full Results}\label{appendix:results}

\begin{figure}
\centering
\includegraphics[width=0.28\textwidth]{figures/results/legend_2.png}\\
\includegraphics[width=\textwidth]{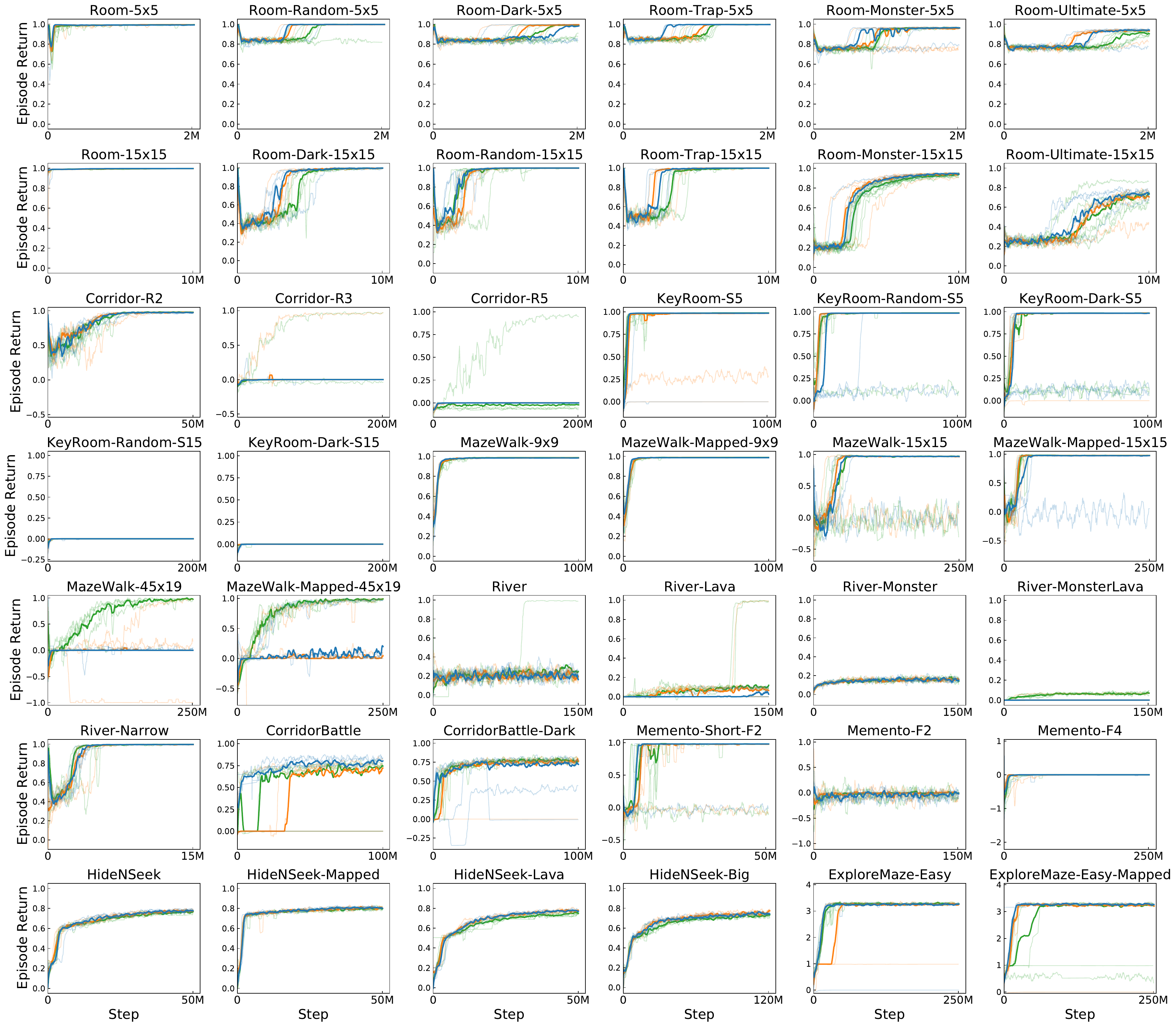}\\
\includegraphics[width=0.35\textwidth]{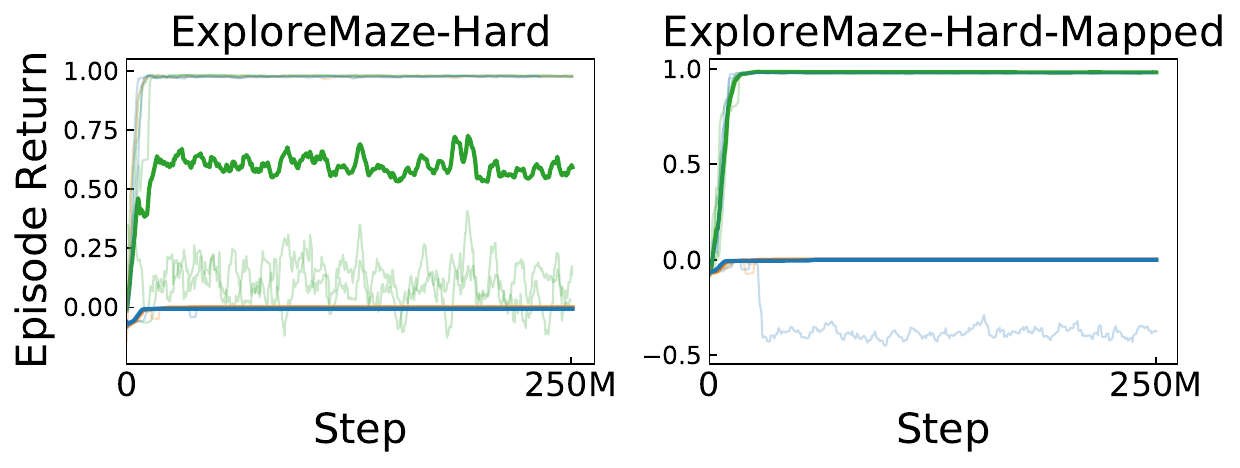}\\
\caption{Mean episode returns on all MiniHack navigation tasks across five independent runs. The median of the runs is bolded.}
\label{fig:res_nav_appendix}
\end{figure}

\begin{figure}
\centering
\includegraphics[width=0.1\textwidth]{figures/results/legend_1.png}\\
\includegraphics[width=\textwidth]{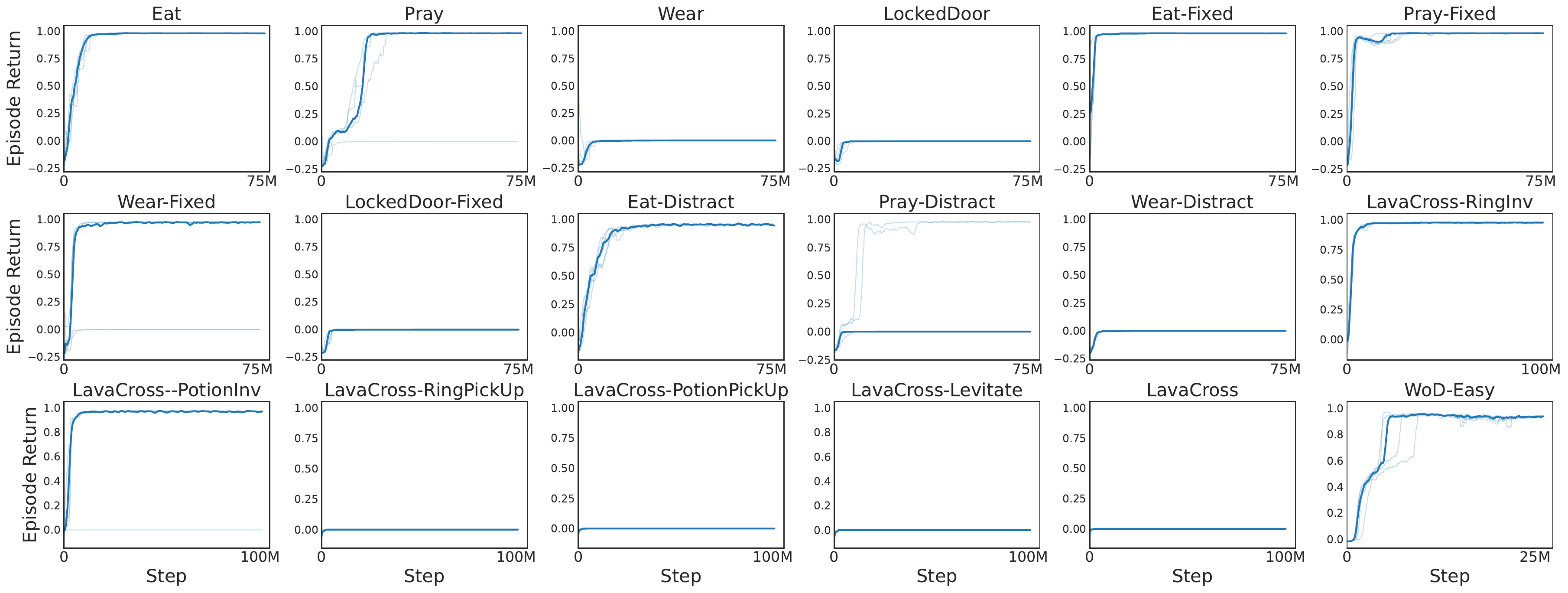}\\
\includegraphics[width=0.9\textwidth]{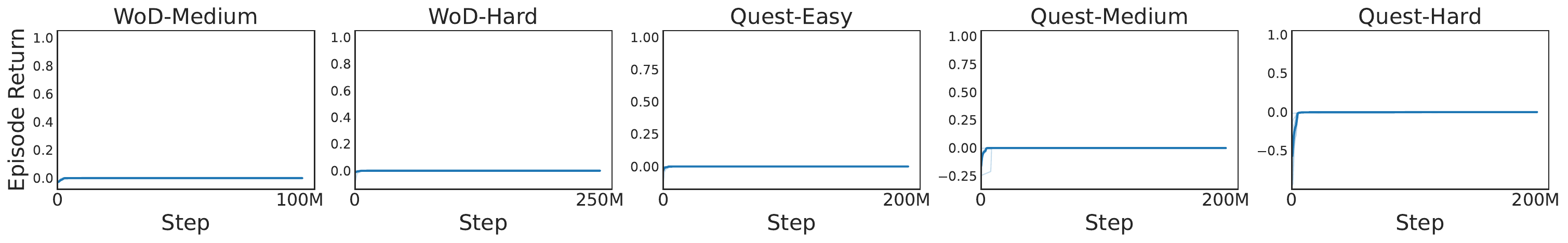}\\
\caption{Mean episode returns on all MiniHack skill acquisition tasks across five independent runs. The median of the runs is bolded.}
\label{fig:res_skill_appendix}
\end{figure}

\begin{figure}
\centering
\includegraphics[width=0.28\textwidth]{figures/results/legend_2.png}\\
\includegraphics[width=\textwidth]{figures/results/port_main.pdf}\\
\includegraphics[width=0.6\textwidth]{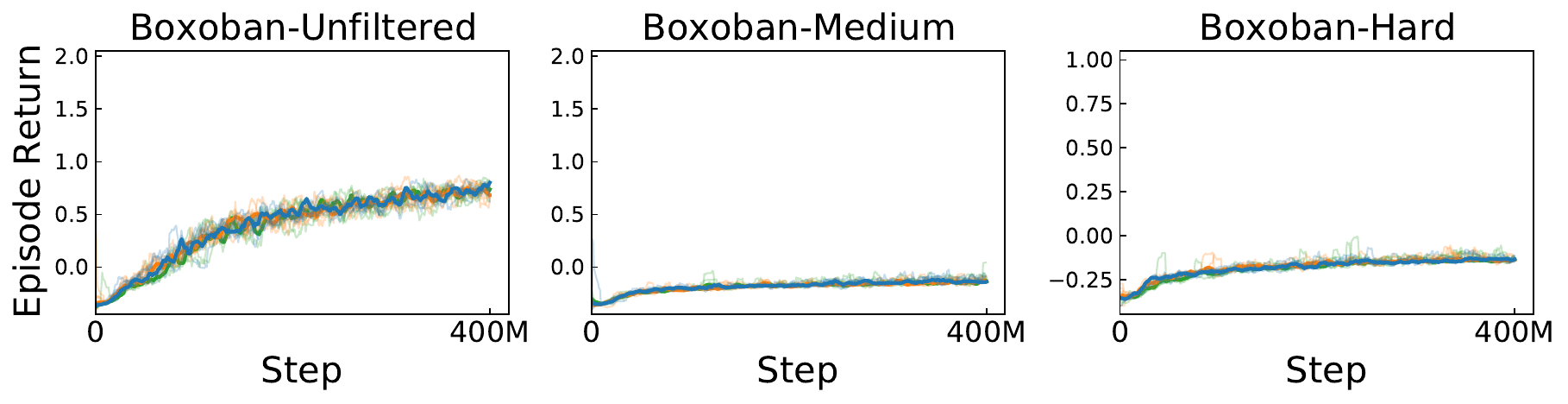}\\
\caption{Mean episode returns on tasks ported to MiniHack from existing benchmarks. The median of the five runs is bolded.}
\label{fig:res_ported_appendix}
\end{figure}

\cref{fig:res_nav_appendix}, \cref{fig:res_skill_appendix}, and \cref{fig:res_ported_appendix} present the results of baseline agents on all navigation, skill acquisition and ported MiniHack tasks, respectively.

\end{document}